\documentclass[10pt,twocolumn,letterpaper]{article}

\usepackage{cvpr}              % To produce the CAMERA-READY version
\usepackage{graphicx}
\usepackage{amsmath}
\usepackage{amssymb}
\usepackage{booktabs}
\usepackage{tabularx,booktabs}
\usepackage{multirow}
\usepackage{mathabx}
\usepackage{adjustbox}
\usepackage{colortbl}
\usepackage{float}
\usepackage[accsupp]{axessibility}

\makeatletter
\@namedef{ver@everyshi.sty}{}
\makeatother
\usepackage{tikz}
\usepackage{etoolbox}
\usepackage{enumitem}
\usepackage{caption}
\usepackage{mwe} % for teaser
\usepackage{makecell}

\newcommand*{\circled}[1]{\lower.7ex\hbox{\tikz\draw (0pt, 0pt)%
    circle (.5em) node {\makebox[1em][c]{\small #1}};}}
\robustify{\circled}

\definecolor{Gray}{gray}{0.92}
\newcolumntype{a}{>{\columncolor{Gray}}c}

\makeatletter
\let\UrlSpecialsOld\UrlSpecials
\def\UrlSpecials{\UrlSpecialsOld\do\/{\Url@slash}\do\_{\Url@underscore}}%
\def\Url@slash{\@ifnextchar/{\kern-.11em\mathchar47\kern-.2em}%
    {\kern-.0em\mathchar47\kern-.08em\penalty\UrlBigBreakPenalty}}
\def\Url@underscore{\nfss@text{\leavevmode \kern.06em\vbox{\hrule\@width.3em}}}
\makeatother

\usepackage[pagebackref,breaklinks,colorlinks]{hyperref}

\usepackage[capitalize]{cleveref}
\crefname{section}{Sec.}{Secs.}
\Crefname{section}{Section}{Sections}
\Crefname{table}{Table}{Tables}
\crefname{table}{Tab.}{Tabs.}

\def\cvprPaperID{1695} % *** Enter the CVPR Paper ID here
\def\confName{CVPR}
\def\confYear{2023}

\begin{document}

\title{One-Shot High-Fidelity Talking-Head Synthesis with Deformable Neural Radiance Field}

%%%%%%%%% ABSTRACT
\twocolumn[{%
\renewcommand\twocolumn[1][]{#1}%
\newcommand*\samethanks[1][\value{footnote}]{\footnotemark[#1]}

\author{Weichuang Li$^1$\thanks{Equal contribution.} \and Longhao Zhang$^2$\samethanks[1] \and Dong Wang$^1$\samethanks[1] \and Bin Zhao$^{1,3}$\thanks{Corresponding author} \and Zhigang Wang$^1$ \and Mulin Chen$^{1, 3}$ \and Bang Zhang$^2$ \and Zhongjian Wang$^2$ \and Liefeng Bo $^2$ \and Xuelong Li $^{1,3}$ 
\samethanks[2]
\vspace{3pt}
\and
$^1$Shanghai AI Laboratory\quad 
$^2$DAMO Academy, Alibaba Group\quad
$^3$Northwestern Polytechnical University\\
% {\tt\small WeichuangLi1999@outlook.com,} \\
{\tt\small\{liweichuang, wangdong, zhaobin\}@pjlab.org.cn}, 
{\tt\small longhao.zlh@alibaba-inc.com}, 
{\tt\small li@nwpu.edu.cn}
}

\vspace{-10mm}

\maketitle
\begin{center}
  \setlength{\belowcaptionskip}{-15pt}
  \centering
  \captionsetup{type=figure, width=.95\textwidth}
  \scriptsize
  \setlength{\tabcolsep}{0pt}
  \adjustbox{max width=0.95\textwidth}{%
  \includegraphics{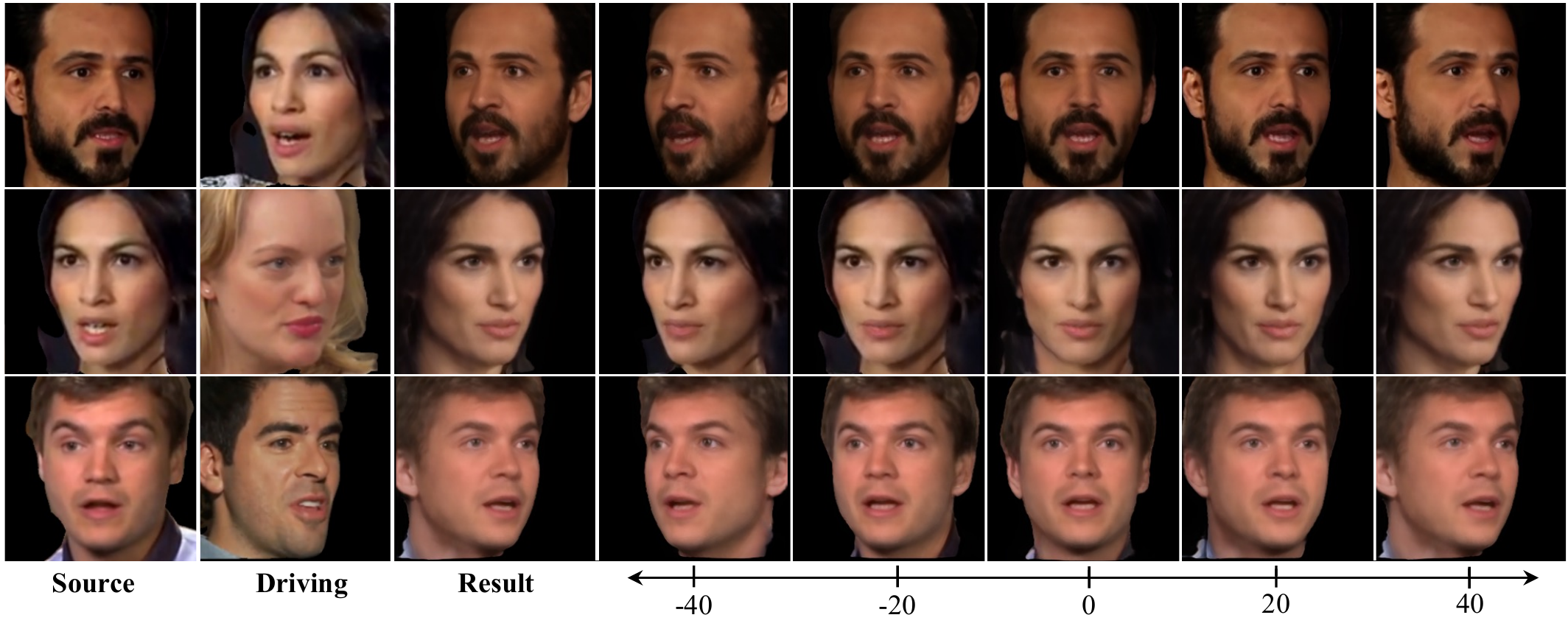}
  }
  \vspace*{-2mm}
%   \captionof{figure}{\textbf{Generated portraits of different poses and expressions.} The first three columns exhibit the source, driving, generated images, respectively. 
% %   The third column exhibits the generated results of our method. 
%   The rest columns show the exploration of the generated images to different yaw angles. }
\captionof{figure}{\textbf{Representative results of our method.} The first three columns exhibit the source, driving, generated images, respectively. 
%   The third column exhibits the generated results of our method. 
  The rest columns show the exploration of the generated images to different yaw angles. }
  \label{fig:teaser}
  \vspace*{4mm}
\end{center}
}]
% \maketitle
\begin{abstract}
Talking head generation aims to generate faces that maintain the identity information of the source image and imitate the motion of the driving image. Most pioneering methods rely primarily on 2D representations and thus will inevitably suffer from face distortion when large head rotations are encountered. Recent works instead employ explicit 3D structural representations or implicit neural rendering to improve performance under large pose changes. Nevertheless, the fidelity of identity and expression is not so desirable, especially for novel-view synthesis. In this paper, we propose HiDe-NeRF, which achieves high-fidelity and free-view talking-head synthesis. Drawing on the recently proposed Deformable Neural Radiance Fields, HiDe-NeRF represents the 3D dynamic scene into a canonical appearance field and an implicit deformation field, where the former comprises the canonical source face and the latter models the driving pose and expression. In particular, we improve fidelity from two aspects: (i) to enhance identity expressiveness, we design a generalized appearance module that leverages multi-scale volume features to preserve face shape and details; (ii) to improve expression preciseness, we propose a lightweight deformation module that explicitly decouples the pose and expression to enable precise expression modeling. Extensive experiments demonstrate that our proposed approach can generate better results than previous works.
Project page: \url{https://www.waytron.net/hidenerf/}
 \end{abstract}
 
\newcommand\blfootnote[1]{%
  \begingroup
  \renewcommand\thefootnote{}\footnote{#1}%
  \addtocounter{footnote}{-1}%
  \endgroup
}

\blfootnote{* denotes equal contribution}
\blfootnote{† denotes the corresponding authors}
% \blfootnote{
% }

%%%%%%%%% BODY TEXT
\baselinestretch
\section{Introduction}
\label{sec:intro}
\baselinestretch

Talking-head synthesis aims to preserve the identity information of the source image and imitate the motion of the driving image. 
Synthesizing talking faces of a given person driven by other speaker is of great importance to various applications, 
such as film production, virtual reality, and digital human. 
Existing talking head methods are not capable of generating high-fidelity results, they cannot precisely preserve the source identity or mimic the driving expression.

Most pioneering approaches~\cite{siarohin2019first, wang2021facevid2vid, hong2022depth, Drobyshev22MP, siarohin2021motion, mallya2022implicit, wang2021facevid2vid}
learn source-to-driving motion to warp the source face to the desired pose and expression. 
According to the warping types, previous works can be roughly divided into: 2D warping-based methods, mesh-based methods, and neural rendering-based methods. 
2D warping-based methods~\cite{siarohin2019first, siarohin2021motion, hong2022depth} warps source feature based on the motion field estimated from the sparse keypoints. 
% These 2D warping-based methods can generate realistic results when the driving image is alike to the source image. 
% For instance, the First-Order-Motion Model~\cite{siarohin2019first} warps source feature based on the motion field estimated from the sparse keypoints.  
% Siarohin \etal~\cite{siarohin2021motion} further extend sparse keypoints into local consistent regions for better accuracy. 
% Recently, DaGAN~\cite{hong2022depth} incorporates the depth estimation to supplement the missing 3D geometry information in keypoints-based motion field. 
However, these methods encounter the collapse of facial structure and expression under large head rotations. 
Moreover, they cannot fully disentangle the motion with identity information of the driving image, resulting in a misguided face shape. 
Mesh-based methods~\cite{grassal2022neural} are proposed to tackle the problem of facial collapse by using 3D Morphable Models (3DMM)~\cite{blanz1999morphable, bfm} to explicitly model the geometry.
% Specifically, ROME~\cite{Khakhulin2022ROME} learns the offset for each mesh vertex and render the rigged mesh with predicted neural texture. 
% Moreover, it introduces a U-Net generator with adversarial loss to refine the rendered images.
Limited by non-rigid deformation modeling ability of 3DMM, such implementation leads to rough and unnatural facial expressions. Besides, it ignores the influence of vertex offset on face shape, resulting in low identity fidelity.
% With the superior capability in multi-view image synthesis of Neural Radiance Fields (NeRF)~\cite{mildenhall2020nerf},
% recent works~\cite{guo2021ad, liu2022semantic, zeng2022fnevr, shen2022dfrf} have been dedicated to directly control head pose of synthetic result with view-dependent volume rendering. 
With the superior capability in multi-view image synthesis of Neural Radiance Fields (NeRF)~\cite{mildenhall2020nerf}, a concurrent work named FNeVR~\cite{zeng2022fnevr} takes the merits of 2D warping and 3D neural rendering. 
It learns a 2D motion field to warp the source face and utilizes volume rendering to refine the warped features to obtain final results. 
Therefore, it inherits the same problem as other warping-based methods.\looseness=-1

To address above issues and improve the fidelity of talking head synthesis, we propose \textbf{Hi}gh-fideltiy and \textbf{De}formable NeRF, dubbed HiDe-NeRF. Drawing on the idea of recently emerged Deformable NeRF~\cite{park2021nerfies, park2021hypernerf, athar2022rignerf}.
HiDe-NeRF represents the 3D dynamic scene into a canonical appearance field and an implicit deformation field. The former is a radiance field of the source face in canonical pose, and the latter learns the backward deformation for each observed point to shift them into the canonical field and retrieve their volume features for neural rendering.
On this basis, we devise a \textit{Multi-scale Generalized Appearance module (MGA)} to ensure identity expressiveness and a \textit{Lightweight Expression-aware Deformation module (LED)} to improve expression preciseness. 

To elaborate, \textit{MGA} encodes the source image into multi-scale canonical volume features, which integrate high-level face shape information and low-level facial details, for better identity preservation.
We employ the tri-plane~\cite{peng2020convolutional, Chan2022} as volume feature representation in this work for two reasons: 
(i) it enables generalization across 3D representations; 
(ii) it is fast and scales efficiently with resolution, facilitating us to build hierarchical feature structures. 
Moreover, we modify the ill-posed tri-plane representation by integrating a camera-to-world feature transformation, so that we can extract the planes from the source image with full control of identity.
This distinguishes our model from those identity-uncontrollable approaches~\cite{Chan2022, bergman2022gnarf} that generate the planes from noise with StyleGAN2-based generators~\cite{karras2020stylegan2}.
Notably,the \textit{MGA} enables our proposed HiDe-NeRF to be implemented in a subject-agnostic manner, breaking the limitation that existing Deformable NeRFs can only be trained for a specific subject.

The deformations in talking-head scenes could be decomposed into the global pose and local expression deformation. The former is rigid and easy to handle, while the latter is non-rigid and difficult to model. 
Existing Deformable NeRFs predict them as a whole, hence failing to capture precise expression.
Instead, our proposed \textit{LED} could explicitly decouple the expression and pose in deformation prediction, thus significantly improving the expression fidelity. Specifically, it uses a pose-agnostic expression encoder and a position encoder to obtain the latent expression embeddings and latent position embeddings, where the former models the expression independently and the latter encodes positions of points sampled from rays under arbitrary observation views. 
Then, a deformation decoder takes the combination of two latent embeddings as input and outputs point-wise deformation. 
In this way, our work achieves precise expression manipulation and maintains expression consistency for free-view rendering (as shown in Fig.~\ref{fig:teaser}).

To summarize, the contributions of our approach are: 
\vspace{-1.5mm}
\begin{itemize}
\setlength\itemsep{-0.25em}

\item Firstly, we introduce the HiDe-NeRF for high-fidelity and free-view talking head synthesis. 
To the best of our knowledge, HiDe-NeRF is the first \textit{one-shot} and \textit{subject-agnostic} Deformable Neural Radiance Fields.

\item Secondly, we propose the \textit{Multi-scale Generalized Appearance module (MGA)} and the \textit{Lightweight Expression-aware Deformation module (LED)} to significantly improve the fidelity of identity and expression in talking-head synthesis. 

\item Lastly, extensive experiments demonstrate that our proposed approach can generate more realistic results than state-of-the-art in terms of capturing the driving motion and preserving the source identity information.
\end{itemize}

\begin{figure*}[t]
  \begin{center}
  
  \includegraphics[width=.95\textwidth]{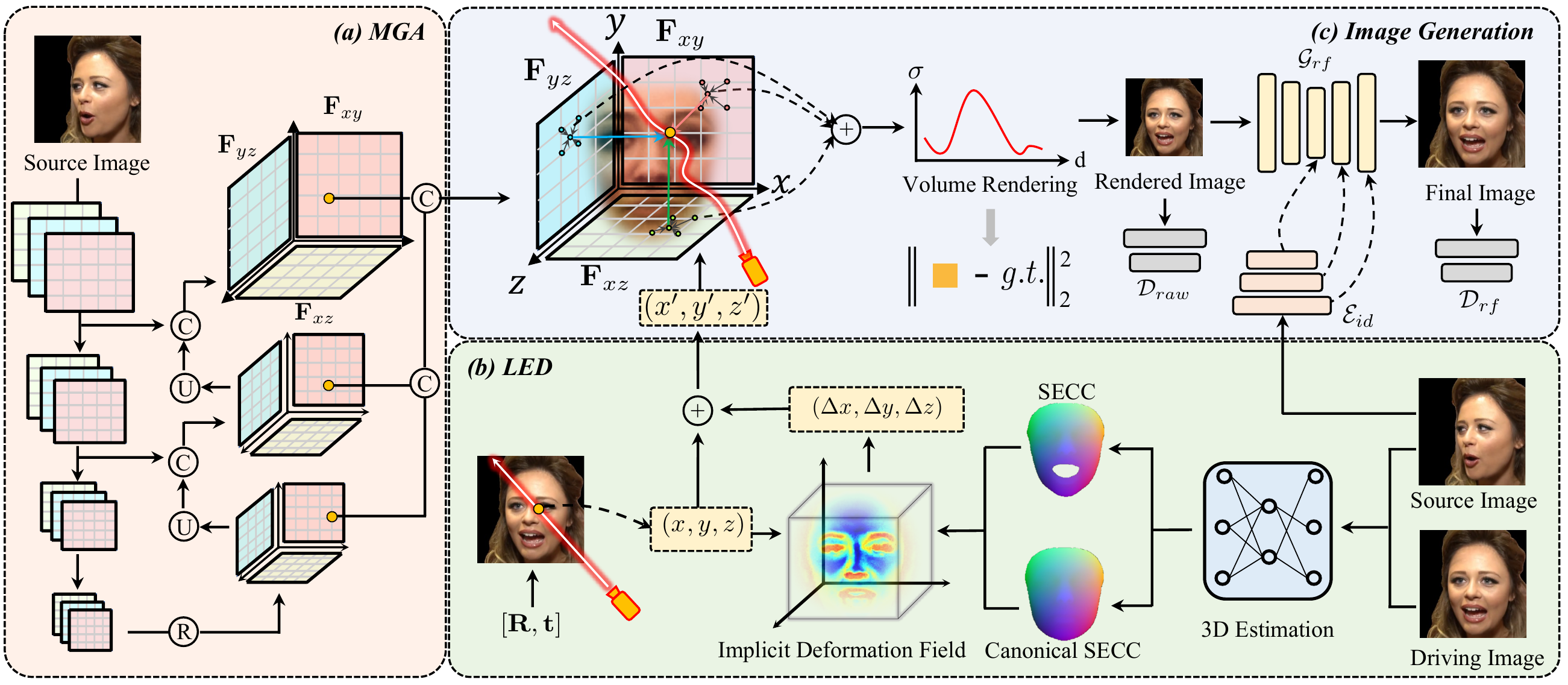}
  \end{center}
  \vspace{-8pt}
  \caption{\textbf{Illustration of the proposed approach. }
  The \textit{MGA} (\textcolor[RGB]{241, 160, 130}{Pink}) encodes the source image into multi-scale canonical volume features. Notably, the skip connection is devised to preserve hierarchical tri-plane properties. 
  The \textit{LED} (\textcolor[RGB]{150, 198, 90}{Green}) predicts backward deformation for each observed point to shift them into the canonical space and retrieve their volume features. The deformation is learned from paired SECCs, conditioned on the positions of points sampled form the rays.
  %point-level offset based on the multi-modal input consists of 3D point positions, driving SECC and the canonical SECC.   
  The image generation module (\textcolor[RGB]{102, 113, 154}{Blue}) takes as input the deformed points to sample features from different scales of tri-planes. These multi-scale features are composed for the following neural rendering.
  %Then, the sampled features and encoded position information are concatenated as input for the NeRF. 
  %In addition to the reconstruction loss, we also implement $\mathbf{D}_{raw}$ as a secondary supervision. 
  %Finally, take the rendered image and source image as input for the $\mathbf{G}_{sr}$ to get the final result. 
  Here we also design a refine network to further refine the texture details (e.g., teeth, skin, hair, etc.), and to enhance the resolution of rendered images.
  Notable, both rendered images and refined images are supervised by reconstruction loss and adversarial loss.
  The symbol {\footnotesize$\circled{\scriptsize{C}}$}, {\footnotesize$\circled{\scriptsize{U}}$},
  {\footnotesize$\circled{\scriptsize{R}}$},
  {\footnotesize$\circled{\scriptsize{+}}$} indicates channel-wise concatenation, upsample, resize and upsample, element-wise sum, respectively. 
  }
  \label{fig:procedure}
  \vspace{-12pt}
\end{figure*}
\baselinestretch
\section{Related Work}
\label{sec:related}
\baselinestretch

\noindent\textbf{Talking-Head Synthesis}.  
% Talking-head synthesis aims to preserve the identity information of the source image and imitate the motion of the driving image. 
Previous talking-head synthesis methods can be divided into warping-based, mesh-based, NeRF-based methods. 
% And we will introduce them accordingly. 
Warping-based methods~\cite{geng2018warp, dong2018soft, liu2019liquid, ha2020marionette, Drobyshev22MP, zhao2022tpsm} are the most popular methods among 2D generation methods~\cite{averbuch2017bringing, burkov2020neural, gu2020flnet, pumarola2018unsupervised,Zakharov20BiLayer}. These methods warp the source features by estimated motion field to transport driving pose and expression into source face. For instance,
Monkey-Net~\cite{Siarohin2019monkey} builds a 2D motion field from the sparse keypoints detected by an unsupervised trained detector. 
% While MonkeyNet assumes a zeroth model, which poorly models appearance transformations. 
% Different from MonkeyNet who assumes a zeroth model, which poorly models appearance transformations, 
% The First-Order-Motion Model~\cite{siarohin2019first} extends Monkey-Net and warps feature based on the local affine transformation learned with the first Taylor expansion of keypoints.
% The First-Order-Motion Model~\cite{siarohin2019first} warps feature based on the local affine transformation learned with the first Taylor expansion of keypoints.
%\wang{Monkey-net pure 2d motion field, DaGAN 2.5D, OSFV fake 3D, focus on warping these three papers}
%To speed up inference, bi-layer~\cite{Zakharov20BiLayer} decomposes images into the high-frequency pose-independent details and low-frequency facial features, and therefore warp the image more efficiently. \wang{delete, no warp in this sentence，replace this with DaGAN, 2D motion field with depth}
DaGAN~\cite{hong2022depth} incorporates the depth estimation to supplement the missing 3D geometry information in 2D motion field.
OSFV~\cite{wang2021facevid2vid} tries to extract 3D appearance features and predict a 3D motion field for free-view synthesis.
%\wang{emphasize 3D motion field}
% DaGAN~\cite{hong2022depth} incorporates the 3D geometry into the motion field generation via a self-learned depth estimator. 
% However, warping-based methods cannot generate realistic image when there is a large angle different between the driving image and the source image. 
% Even though ~\cite{wang2021facevid2vid, hong2022depth} tries to leverage the 3D information, the occlusion parts are generated through image inpaiting. 
% Another disadvantage of warping-based models is that they are limited in shape preservation since key points contain both shape and expression information. 
% \noindent\textbf{Mesh-based talking-head  synthesis. } 
Some conventional works~\cite{suwajanakorn2017synthesizing, thies2019deferred, thies2015real, thies2016face2face} employ 3D Morphable Models (3DMM)~\cite{blanz1999morphable, bfm}, which support a diverse range of animations via disentangled shape, expression and rigid motions.
StyleRig~\cite{tewari2020stylerig} and PIE~\cite{tewari2020pie} are proposed to exploit the semantic information in the latent space of StyleGAN~\cite{karras2019stylegan} and modulate the expressions using 3DMM. 
% PIE further improves the fidelity by incorporating an optimization-based inversion method to calculate
% the optimal latent code. 
PIRender~\cite{ren2021pirenderer} uses 3DMM to predict the flow and warp the source image. 
% Face2Face$\rho$ ~\cite{yangface2face} also designs a 3DMM-assisted warping-based model, it uses different scales of landmark images to predict facial motion fields. 
ROME~\cite{Khakhulin2022ROME} is the first 3DMM-based method that uses a single image to create realistic photo in a rigged mesh format. 
It learns the offset for each mesh vertex and renders the rigged mesh with predicted neural texture. 
Moreover, it introduces a U-Net generator with adversarial loss to refine the rendered images. 
% \noindent\textbf{NeRF-based talking-head  synthesis}. 
We also use 3DMM in this work, but unlike other approaches, we use it to generate the Shape- and Expression-aware Coordinate Code ($\mathbf{SECC}$) to build our deformation field, which is detailed in Sec.~\ref{sec:uvdeform}. 
NeRF is a recently proposed implicit 3D scene representation method that renders the static scene with points along different view directions. 
% Recently, there are also some works~\cite{park2021hypernerf, park2021nerfies} extend NeRF into dynamic scenes. 
% NeRFPlayer~\cite{song2022nerfplayer}, 
NeRF first flourished in the audio-driven approaches~\cite{guo2021adnerf, yao2022dfa, liu2022semantic, shen2022dfrf}, as they can easily combined with the latent code learned from the audio. 
But they are restricted to subject-dependent generation. 
% HeadNeRF~\cite{hong2021headnerf} uses the parameter codes estimated with 3DMM, and concatenates the codes with positional encoded points as input to the NeRF. 
% FNeVR~\cite{zeng2022fnevr} is related to our approach in that it is also a subject-agnostic method. 
% Differently, it relies on the accurate expression manipulation of warping-based methods. 
% It first warp the image according to the motion field estimated with keypoints, and then feed the warped feature into NeRF to enhance the facial details. 
% The generated image is obtained from the concatenation of warped and rendered feature maps. 

% \vspace{-0.2cm}
\noindent\textbf{Deformable Neural Radiance Field. }
Deformable NeRF is proposed for rendering of dynamic scene. NeRFies~\cite{park2021nerfies} maps each observed point into a canonical space through a continuous deformation field represented by a scene-specific MLP, and it can only handle small non-rigid movements.
HyperNeRF~\cite{park2021hypernerf} inherits NeRFies and uses an ambient dimension to model the topological changes in the deformation field. 
%The deformation field in these two methods is conditioned on learned latent codes. 
Conversely, RigNeRF~\cite{athar2022rignerf} uses 3DMM to learn the deformation by finding the closest driving mesh vertex for each sampled point and explicitly calculating its distance from the corresponding canonical mesh vertex, which is not efficient or precise enough.
Notably, all the above-mentioned Deformable NeRFs are subject-dependent, which implies they need to train an individual model for each subject. Besides, they treat the global pose and local expression deformation as a whole, thus cannot achieve precise expression manipulation.
In this work, we draw on the idea of Deformable NeRF and propose the HiDe-NeRF.
Different from existing methods, our approach is an \textit{one-shot} and \textit{subject-agnostic} Deformable NeRF specially designed for talking head synthesis. 
Besides, our \textit{LED}-based deformation field is computationally much more efficient than other MLP-based deformation fields.\looseness=-1

% \clearpage
% \vspace{-0.1cm}
\section{Methods}
% \vspace{-0.1cm}

%We present High-fidelity and Deformable NeRF, dubbed HiDe-NeRF, that generates id-controllable and high-fidelity portraits. 
In this section, we describe our method, HiDe-NeRF, that enables high-fidelity talking head synthesis.
% The overall procedure can be illustrated in Fig.~\ref{fig:procedure} and we detail on them in Sec.~\ref{sec:overview}.
The overall procedure can be illustrated in Fig.~\ref{fig:procedure}.
% We will detail on the overall procedure in  
%Then, we will review the tri-plane representation and analyze its limitations for conditional generation, and introduce the \textit{Appearance Module} that comprises delicate identity information (Sec.~\ref{sec:tri-plane}).  
%Afterwards, we expound the \textit{ Deformation module} that handles the expression changes (also called non-rigid deformations)  (Sec.~\ref{sec:uvdeform}). 
%Then, we will describe the image generation module, which is composed of volume rendering and refine module (Sec.~\ref{sec:nerf}). 
% handles pose changes, and the supper resolution network. 
Then, we expound the proposed \textit{Multi-scale Generalized Appearance module (MGA)} and \textit{Lightweight Expression-aware Deformation module (LED)} in Sec.~\ref{sec:tri-plane} and Sec.~\ref{sec:uvdeform}.
Afterward, we describe the image generation module in Sec.~\ref{sec:nerf}, including volume rendering and texture refinement.
The training details can be found in the Supplementary Materials. 
%Finally, the training details are displayed in Sec.~\ref{sec:detail}.
\looseness=-1

\subsection{Multi-Scale Generalized Appearance Module}\label{sec:tri-plane}
% \vspace{-0.1cm}

% \input{Figs/triplane}
% \wang{I think, contribution of this section: 1.multi-scale tri-plane to improve synthesis quality in this one-shot setting(talking head); 2.camera-to-world makes learning much better. the first one is unclear.}

As discussed, Deformable NeRF~\cite{park2021nerfies, park2021hypernerf, athar2022rignerf} typically targets novel view synthesis based on multi-view inputs under a single-scene scenario. 
Different from the settings of conventional Deformable NeRF, talking-head synthesis aims to generate high-fidelity images of any particular person with a single image, which can be summarized as subject-agnostic and high-fidelity preservation under one-shot setting.

In this work, we introduce tri-plane representation as our appearance field to accommodate the attribute of subject-agnostic. 
Tri-plane hybrid representation~\cite{Chan2022, peng2020convolutional} is recently proposed, which builds three orthogonal planes with feature maps. 
% As shown in Fig.~\ref{fig:encoding}(a),
Given a 3D point $\boldsymbol{p}=(x,y,z)$, it is projected onto $\mathbf{F}_{xy}, \mathbf{F}_{xz}, \mathbf{F}_{yz}$ three planes to query the feature vectors via bilinear interpolation. 
The queried features from three planes are averaged as the representation of point $F(\boldsymbol{p}) = (F_{xy}(\boldsymbol{p}) + F_{xz}(\boldsymbol{p}) + F_{yz}(\boldsymbol{p})) / 3$, where $F_{ij} : \mathbb{R}^3 \mapsto \mathbb{R}^C$ denotes sampling the feature of 3D coordinates from the planar feature map $\mathbf{F}_{ij}$.  
%To achieve ID-controllable generation, the tri-plane representation was constructed with the volume features extracted from the source image. 
% \textbf{Camera-to-world transformation makes ID-controllable tri-plane learning much better. }
% Recent literature applies tri-plane representation into the  unconditional generation of articulated objects~\cite{bergman2022gnarf}, and 3D shapes~\cite{wu2022learning}. 
Although the tri-plane representation has been applied in different areas~\cite{bergman2022gnarf, zhu2022nice, DeepFaceVideoEditing2022}, there is no method that directly extract planes from an image. 
We find one core issue in learning tri-plane representation from the image is that the camera coordinate system and the world coordinate system are oriented in different directions. 
The definition of tri-plane is based on the world coordinate system, but the axes of the image are aligned with the camera coordinate system. 
Due to this problem, the predicted volume features $\{\mathbf{V}_i, i=1,2,3\}$, where $\mathbf{V}_i \in \mathbb{R}^{c \times h \times w}$ from deep network are mismatched with the definition of tri-plane representation, which makes the representation difficult to learn from the image directly. 
Therefore, we use source camera parameters $\{\mathbf{R}_{src}, \mathbf{t}_{src}\}$ to transfer the predicted volume features into tri-plane representation.
% as shown in Fig.~\ref{fig:encoding}(b). 
%This schema enables us to extract features from source image directly. 
Concretely, this transformation could be formulated as below,
\vspace{-3mm}
\begin{equation}
\label{eq-c2w}
\begin{gathered}
\mathbf{F}_{plane}=\mathcal{T}\left(\left\{ \mathbf{R}_{src}, \mathbf{t}_{src}\right\}, \mathbf{V}_i\right), i=1,2,3, \\
% \mathbf{F}_{xz}^{trans}=\mathcal{T}\left(\left\{ \mathbf{R}_{src}, \mathbf{t}_{src}\right\},\left\{x, 0, z\right\}, \mathbf{F}_{xz}\right), \\
% \mathbf{F}_{yz}^{trans}=\mathcal{T}\left(\left\{ \mathbf{R}_{src}, \mathbf{t}_{src}\right\},\left\{0, y, z\right\}, \mathbf{F}_{yz}\right),
\end{gathered}
\end{equation}
where $\mathcal{T}$ denotes the camera-to-world transformation function and $plane \in [xy, xz, yz]$.

Another challenging issue of talking-head generation is preserving identity fidelity under the one-shot setting. 
% The expressiveness of tri-plane representation is determined by the feature maps that were used for its construction, 
% but the information capability of single-scale feature maps is usually limited. 
% The vanilla tri-plane representation suffers from two issues: (i) the information capability is limited;  (ii) the learning of intermediate feature map is under-supervised. 
It is known that different scales of feature provide different information, high-level feature maps comprise facial shape information while low-level feature maps contain facial details, for instance, skin texture, makeup, \etc. 
To improve the expressiveness of tri-plane representation, we adopt multi-scale tri-plane representation instead, which integrates different levels of semantic information. 
As shown in Fig.~\ref{fig:procedure}(a), we first employ a deep feature extractor to derive the pyramid feature maps  $[\mathbf{M}^{0},\dots ,\mathbf{M}^{n}]$ from the source image $I_{src}$. 
For the lowest-resolution feature map $\mathbf{M}^{0}$, a small convolutional decoder ${\psi}^0$ is used to predict volume features $\mathbf{V}^0$, and the corresponding lowest-scale tri-plane representation $\mathbf{F}^{0}$ is obtained by applying camera-to-world transformation in Eq.~\ref{eq-c2w}.
Based on this, the multi-scale tri-plane representations are formulated as: 
\begin{equation}
\vspace{-1mm}
\begin{gathered}
\mathbf{V}^{j+1}_k = \mathcal{\psi}^{j+1}_k([\mathbf{M}^{j+1}_k, \mathbf{F}^j_k\uparrow]), \\
\mathbf{F}^{j+1}_k = \mathcal{T}\left(\left\{ \mathbf{R}_{src}, \mathbf{t}_{src}\right\}, \mathbf{V}^{j+1}_k\right),
\end{gathered}
\vspace{-1mm}
\end{equation}
where $k \in \{1,2,3\}$ represent different planes, $\mathcal{\psi}_k^{j+1}$ denotes a convolutional network, $\uparrow$ denotes the up-sampling operation. $\mathbf{V}_k^j$ is the $j$-th level of the predicted volume feature. 
% , where $\mathbf{F}_{tri}^{n} \in \mathbb{R}^{(3 \times c) \times h \times w}$ is a tri-plane volume feature. 
% The sampled representation of single-scale tri-plane representation is composed of local low-level information. 
Based on this multi-scale tri-plane representation, the final representation of a single point $\boldsymbol{p}$ can be formulated as $F(\boldsymbol{p}) = [F^{0}(\boldsymbol{p}), \dots F^{n}(\boldsymbol{p})]$, where $[\dots]$ denotes channel-wise concatenation.
\subsection{Lightweight Expression-Aware Deformation Module}\label{sec:uvdeform}
% \vspace{-0.1cm}

%\wang{I think, highlights of this section, SECC enable model precise expression; point-wise deformation for free-view synthesis.}

\begin{figure}[tb]
  \centering
  \includegraphics[width=\linewidth]{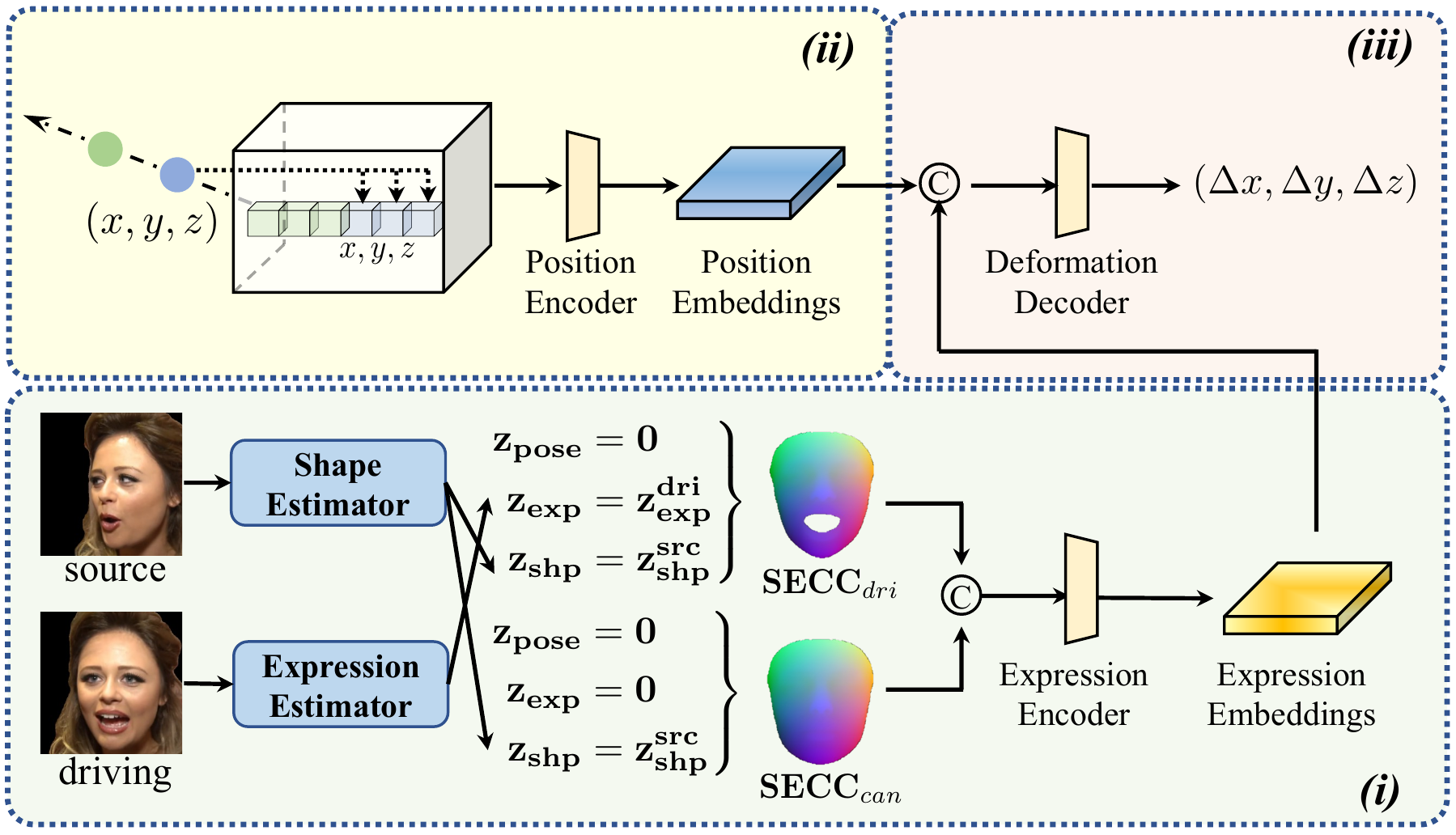}
  \vspace{-5mm}
  \caption{\textbf{The illustration of the proposed \textit{LED}.} 
  The symbol {\footnotesize$\circled{\scriptsize{c}}$} indicates channel-wise concatenation.  
  \looseness=-1
  }
  \label{fig:uvdeform}
  \vspace*{-8pt}
\end{figure}
% Talking head generation comprises multiple head motions, person appearances, dynamic expressions, which is difficult for NeRF network to learn. 
% Defining a canonical space will alleviate the learning difficulty 
The deformations in talking-head scenes could be decomposed into the global rigid pose and local non-rigid expression deformation. 
Existing Deformable NeRFs predict them as a whole, hence failing to accurately model complex and delicate expressions.
%Existing Deformable NeRFs model them by a whole deformation field, hence resulting in insufficient capability to model precise expression.
%\zlh{we are also "a whole deformation field". We encode expressions and pts separately in this field}
In this work, we propose the \textit{ Lightweight Expression-aware Deformation Module (LED)}, which explicitly decouples the expression and pose in deformation prediction, significantly improving the expression fidelity.
Moreover, the decoupling of expression and pose ensures expression consistency for free-view rendering.
As illustrated in Fig.~\ref{fig:uvdeform}, the \textit{LED} could be divided into three steps:\looseness=-1 

%\noindent\textbf{\textit{(ii) Represent the expression deformation.}}
\noindent\textbf{\textit{(i) Expression Encoding.}}
First, we introduce the Shape- and Expression-aware Coordinate Code (SECC) to learn the pose-agnostic expression deformation for precise expression manipulation.
SECC is obtained by rendering a 3DMM face ~\cite{bfm} through Z-Buffer with Normalized Coordinate Code (NCC) ~\cite{zhu2017face} as its colormap. It could be formulated as: 
\vspace{-1mm}
\begin{equation}\label{eq:secc}
\begin{gathered}
\mathbf{SECC}=\operatorname{Z-Buffer}\left(V_{3 d}(\mathbf{p}), \mathbf{NCC}\right), \\
V_{3 d}(\mathbf{p}) = \mathbf{R}\left(\overline{\mathbf{S}}+\mathbf{A}_{shp} \boldsymbol{z}_{shp} +\mathbf{A}_{exp} \boldsymbol{z}_{exp}\right) + \mathbf{t},
\end{gathered}
\vspace{-2mm}
\end{equation}
where $\overline{\mathbf{S}}$ is the template shape, $\mathbf{A}_{shp}$ and $\mathbf{A}_{exp}$ are principle axes for shape and expression. We set $\mathbf{R} = \mathbf{1}$ and $\mathbf{t}= \mathbf{0}$ to eliminate the pose.

As shown in Fig.~\ref{fig:uvdeform}(i), we use a pair of SECC to model the shape-aware expression changes from driving to canonical.
Specifically, we predict the expression coefficient  $\mathbf{z}_{exp}^{dri}$ from the driving image  $\mathbf{I}_{dri}$ and the shape coefficient  $\mathbf{z}_{shp}^{src}$ from the source image  $\mathbf{I}_{src}$ using the 3D Estimator\cite{deng2019accurate}, and we form the driving $\mathbf{SECC}_{dri}\in \mathbb{R}^{H\times W \times 3}$  with source shape $\mathbf{z}_{shp}^{src}$ and driving expression $\mathbf{z}_{exp}^{dri}$,  and the canonical $\mathbf{SECC}_{can}\in \mathbb{R}^{H\times W \times 3}$ with source shape $\mathbf{z}_{shp}^{src}$ and zero expression $\mathbf{z}_{exp}^{0}$.
Since the $rgb$ value of each point in NCC ~\cite{zhu2017face} corresponds to the $xyz$ coordinate of a specific mesh vertex, it establishes the vertex-to-pixel correspondence between 3D and 2D. Therefore, we directly apply a 2D convolutional encoder on paired SECCs to learn the latent expression embeddings that contains 3D expression deformation.\looseness=-1

%\noindent\textbf{\textit{(i) Represent the 3D scene.}} 
\noindent\textbf{\textit{(ii) Position Encoding. }} 
%\wang{Render Pose Encoding}
%For a 3D scene with $U\times V \times N$ points, where $N$ indicates how many samples along each ray and $U$ and $V$ denote the rendering resolution. 
%We rearrange the 3D volume points according to their positions. 
%Concretely, we regard points that share the same depth as the same level and arrange points that belong to the same ray at the same 2D position.  
%Then, we separate their coordinates into three channels (x, y, z, accordingly) to represent the positional information.  
%Finally, the 3D scene can be represented with $\mathbf{P \in \mathbb{R}^{U\times V \times (3 \times N)}}$.
In order to learn the point-wise deformation under the observation view (can be the driving image view or arbitrary views), we encode the 3D coordinate of points sampled from the rays as the positional condition. 
%\wang{as another input of LED module}.
%\wang{To this end logic is unclear, no aim is illustrated before}
Specifically, we first reshape the points $\mathbf{P} \in \mathbb{R}^{H \times W \times N \times 3}$ to $\mathbf{P} \in \mathbb{R}^{H\times W \times (3 \times N)}$,  where $N$ is the number of sampled points along each ray, $H$ and $W$ denote the rendering resolution.
It is then fed into a fully-convolutional position encoder to get the latent position embeddings.
% Concretely, we rearrange the 3D volume points back to the UV space and separate the corresponding coordinates into three channels to represent the positional information.

\noindent\textbf{\textit{(iii) Deformation Prediction.}}
%\wang{emphasize our deformation only model expression? or capability of free-view synthesis}
The latent expression embeddings and the latent position embeddings are concatenated in channel-wise and fed into a deformation decoder to predict the point-level deformation $\Delta \mathbf{P} \in \mathbb{R}^{(H\times W) \times (3\times N)}$.

In summary, for a deformation module parameterized by $\Phi$, the implicit function can be formulated as:

\vspace{-5mm}
\begin{equation}
\mathcal{F}_{\Phi}^{\text {deform }}:\left(\mathbf{P}, \mathbf{SECC}_{dri}, \mathbf{SECC}_{can} \right) \rightarrow \Delta \mathbf{P}.
\vspace{-2mm}
\end{equation}

As mentioned above, our proposed \textit{LED} employs the vertex-to-pixel correspondence and the positional encoding to learn point-wise 3D deformations. It is lightweight yet efficient since it doesn't need to find the closest driving mesh vertex for each sampled point and explicitly calculate its distance from the corresponding canonical mesh vertex like ~\cite{athar2022rignerf}. Besides, the encoder and decoder network in \textit{LED} are fully-convolutional and very shallow, thus is computationally much more efficient than other MLP-based deformation fields~\cite{park2021nerfies, park2021hypernerf, athar2022rignerf}.

\subsection{Image Generation Module}\label{sec:nerf}
% \vspace{-0.1cm}

The image generation module is composed of volume rendering and texture refinement. 

\noindent\textbf{Volume Rendering.}
Given camera intrinsic parameters and camera pose of driving image, we calculate the view direction $\mathbf{d}$ of a pixel coordinate $(h,w)$. We first sample $N$ points along this ray for stratified sampling. Formally, let $\mathbf{p}_i = \mathbf{o} + t_i\mathbf{d}, i \in \{1,...,N\}$ denote the sampling points on the ray given camera origin $\mathbf{o}$, and $t_i$ corresponds to the depth value of $\mathbf{p}_i$ along this ray. For every point $\mathbf{p}_i$, we first apply positional encoding $\gamma(q)=<\sin(2^l\pi q), \cos(2^l\pi q)>$ to it, and sample volume feature ${F}(\mathbf{p}_i^{\prime})$ from multi-scale tri-planes at deformed point position $\mathbf{p}_i^{\prime} = \mathbf{p}_i + \mathbf{\Delta p}_i$, where $\mathbf{\Delta p} \in \mathbb{R}^{3\times N}$ is the $(h\times H + w)$-th row vector of $\Delta \mathbf{P}$. Then ${F}(\mathbf{p}_i^{\prime})$ and $\gamma(\mathbf{p}_i)$ are concatenated as $\mathbf{f}(\mathbf{p}_i) =[F(\mathbf{p}_i^{\prime}), \gamma(\mathbf{p}_i)]$ and fed into a two-layer MLP to predict color $\mathbf{c}$ and density $\mathbf{\sigma}$ of point $\mathbf{p}_i$. Finally, the color of this pixel can be rendered as:

\vspace{-3mm}
\begin{equation}
\begin{aligned}
\mathbf{C}(\mathbf{r})&=\int_{t_n}^{t_f} T(t) \sigma(\mathbf{r}(t)) \mathbf{f}(\mathbf{r}(t)) \mathrm{d} t, \\
\quad T(t)&=\exp \left(-\int_{t_n}^{t_f} \sigma(\mathbf{r}(s)) \mathrm{d} s\right),
\end{aligned}
\vspace{-0.1mm}
\end{equation}
where $t_n$ and $t_f$ indicate near and far bounds along the ray. %$\mathbf{r}(t) = \mathbf{o} + t\mathbf{d}$. 

\noindent\textbf{Texture Refinement.} 
%Since the rendered image does not contain much texture detail, first, we use an identity extractor $\mathcal{E}_{id}$ to extract the identity information from the source image. 
Since the texture details (e.g., teeth, skin, hair, etc.) and the resolution of the rendered images $\mathbf{I_{raw}}$ are limited, we design an refine network $\mathcal{G}_{rf}$ with encoder-decoder structure to improve them and generate the final image $\mathbf{I_{rf}}$.
Specifically, we use an identity extractor $\mathcal{E}_{id}$ to extract multi-scale texture features from the source image, and inject them to the decoder of $\mathcal{G}_{rf}$ through SPADE~\cite{park2019semantic}. 
Notably, $\mathbf{I_{rf}}$ and $\mathbf{I_{raw}}$ are fed into two separate discriminators for adversarial training.
Details of the refine network are illustrated in the supplementary.
\begin{figure*}[ht]
  \centering
  \includegraphics[width=0.95\linewidth]{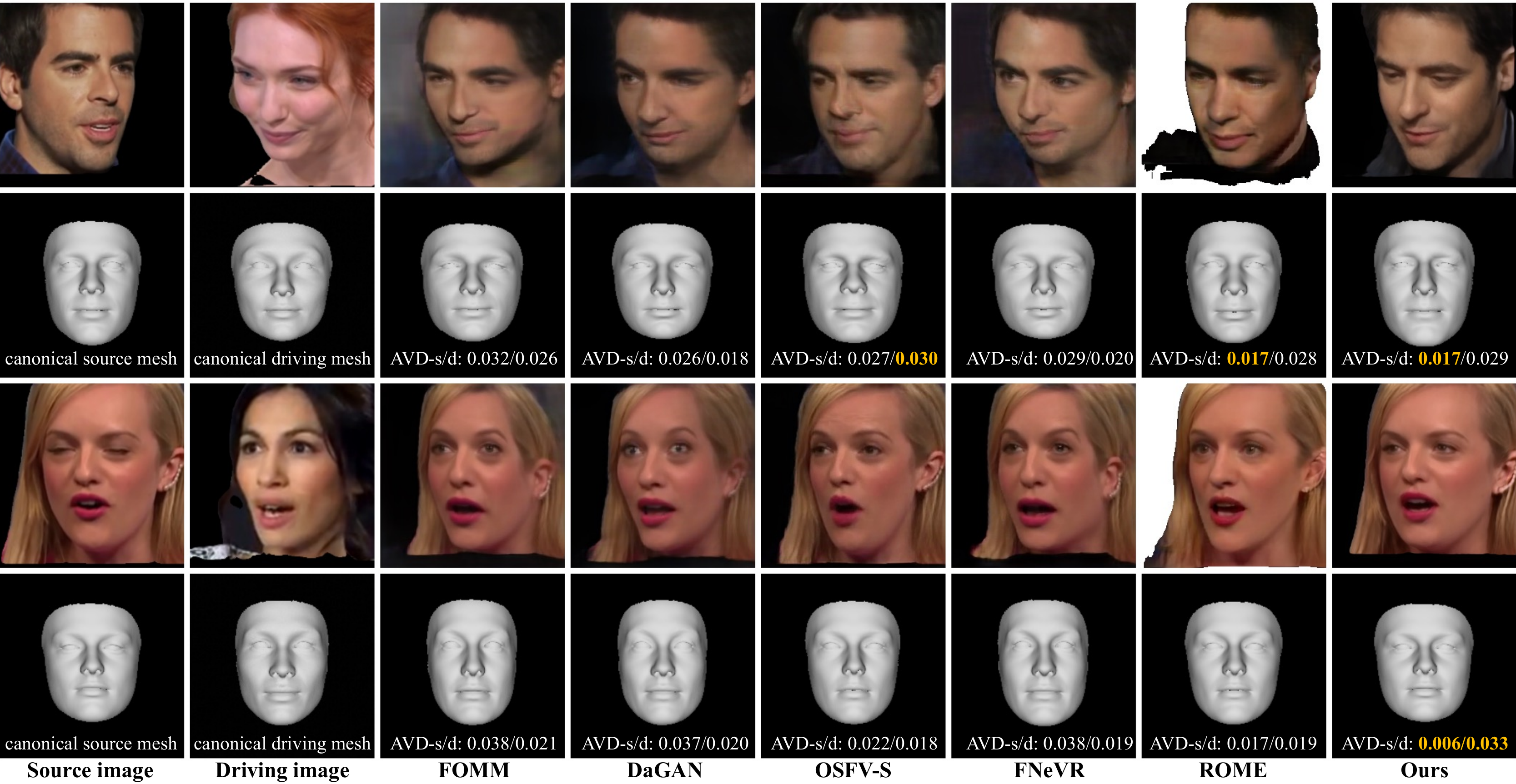}
  \vspace{-3pt}
  \caption{
\textbf{Comparison of shape preservation with prior works. } 
AVD-s/d indicate average vertices distance with source and driving mesh. 
The first and third row contains the source, driving, and generated images. 
The second and fourth row includes the corresponding canonical mesh of their above image.  
Best results are marked with yellow. 
 }
  \label{fig:shape_com}
  \vspace{-12pt}
\end{figure*}

% \vspace{-0.1cm}
\section{Experiments and Results}
% \vspace{-0.1cm}
\subsection{Dataset Preprocessing and Metrics}
% \vspace{-0.1cm}

\noindent\textbf{Dataset Preprocessing.}
% We take VoxCeleb1~\cite{Nagrani17} as the training dataset and use Deep3DFace~\cite{deng2019accurate} to align face and calculate the proposed SECC. 
% We take  as the training dataset and test the generalization capability of our model on VoxCeleb1 and . 
We conduct experiments over three commonly used talking-head generation datasets (\ie VoxCeleb1~\cite{Nagrani17}, VoxCeleb2~\cite{Chung18b}, TalkingHead-1KH~\cite{wang2021facevid2vid}). 
Each frame is cropped and aligned into $256\times256$ to center the talking portraits, 
their rotation angles are predicted with Face-Alignment~\cite{bulat2017far}. 
%Please note that this pre-processing procedure induces larger facial parts, which makes the evaluation metrics more demanding in terms of facial details. 

% \wang{flip is a data-augmentation, re-write this paragraph from data-augmentation perspective}
% NeRF is known to benefit from different view inputs, but there are often subtle pose changes  during short video clips.
% Therefore, in the following experiments, similar to the conventional 3D GANs~\cite{Chan2022}, we set the flipping probability $p_{flip}=0.5$.
% Nevertheless, flipping leads to changes in the background.
% % , resulting in a significant loss of MSE.
% Therefore, we only preserve the head and torso parts in the video frames via an off-the-shelf segmentation predictor~\cite{face-parsing}.\looseness=-1

\noindent\textbf{Metrics.} \hspace{-1.5mm}
We measure the quality of synthetic images using structured similarity (\textbf{SSIM}), \textbf{PSNR} (masked, only compare the region of face, hair, and torso),  \textbf{LPIPS}~\cite{zhang2018unreasonable}, and \textbf{FID}. 
% Also, we choose \textbf{LPIPS}~\cite{zhang2018unreasonable} to measure the fidelity. 
Following the previous work~\cite{ha2020marionette, hong2022depth}, we adopt fidelity metrics \textbf{CSIM}, \textbf{PRMSE}, \textbf{AUCON} to evaluate the identity preservation of the source image, the accuracy of head poses and the precision of expression, respectively. 

Furthermore, we propose a new metric named average vertices distance (\textbf{AVD}) for better identity preservation evaluation. 
%In particular, we first obtain the 3D reconstructed face with ~\cite{deng2019accurate} and transform its mesh into a canonical space to exclude the influence of pose and expression. 
%Then, we calculate the distances between mesh vertices and take their average value as the final metric, termed as AVD.
% In particular, we first obtain the face meshes with~\cite{deng2019accurate}, and set the pose and expression coefficients to 0 to eliminate their impact on face shape.
To do this, we first obtain face meshes using~\cite{deng2019accurate} and neutralize the impact of pose and expression by setting the corresponding coefficients to 0. 
% In particular, we first obtain the face meshes with~\cite{deng2019accurate}, and eliminate their pose and expression.
Then we calculate the mesh vertices distance between generated face and source face as AVD-s, and that between synthesized face and driving face as AVD-d.
% Then we calculate the mesh vertices distance between generated face and source face as AVD-s, and that between synthesized face and driving face as AVD-d.
% As shown in Fig.~\ref{fig:shape_com}, 
% compared with other methods, our approach has a lower AVD-s and a higher AVD-d, which means we can well preserve the source face shape, and not be misguided by the driving face shape. 
As Fig.~\ref{fig:shape_com} illustrates, our approach outperforms other methods by exhibiting a lower AVD-s and a higher AVD-d, which indicates that we can preserve the source face shape effectively and not be swayed by the driving face shape.
%lower AVD-s indicates that the generated face retains the source face shape better. 
% shape distance \wang{AVD} to the source image 
%We also report the AVD \wrt the driving image (AVD-d) in Fig.~\ref{fig:shape_com}. 
%Notably, a higher AVD-d only implies that the generated face is less influenced by the driving face, rather than having a better shape fidelity. 
% to the shape of the driver image only implies that the generated shape is less influenced by the driving shape, \wang{rather than better shape fidelity}. but does not necessarily indicate better shape fidelity. 
% Since AVD-d does not reflect identity preservation, in the following parts, we only report AVD-s, and abbreviate it as AVD. 
Since AVD-d does not reflect identity preservation, we only report AVD-s and abbreviate it as AVD in future discussions.
%It can be observed that for warping-based methods the shape distance between the driving image is less than it is with the source image, which validates that the synthetic face shape of warping-based methods are affected by the driving image.

% OSFV~\cite{wang2021facevid2vid} disentangle the identity-related information from its keypoint representations in an unsupervised manner, therefore, has similar distances to the source and driving (0.027/0.030 and 0.022/0.018). 
% Our method achieve the lowest \wang{no AVD explanation in previous text}AVD \wrt the source image, which demonstrates our method preserve the source shape with the highest fidelity. 
%Our method achieves the lowest AVD \wrt the source image, which demonstrates our method preserve the source shape with the highest fidelity. 

% \vspace{-0.1cm}
\subsection{Talking-Head Synthesis}
% \wang{Comparison with State-of-the-art Methods?}
% \vspace{-0.1cm}

\noindent\textbf{Baselines.}
We compare our method against five state-of-the-art methods, including 2D-warping-based methods: FOMM~\cite{siarohin2019first}, OSFV-S~\cite{wang2021facevid2vid} (``-S" indicates the model with SPADE~\cite{park2019semantic} layers, which produces better results), DaGAN~\cite{hong2022depth} and 3D-based methods: ROME~\cite{Khakhulin2022ROME} (Mesh-based), FNeVR~\cite{zeng2022fnevr} (NeRF-based). 
All results are obtained by evaluating these method with their official code.\looseness=-1 
% \lwc{We use the official released model for ROME and FNeVR and train other methods on VoxCeleb1. }

\begin{table}[t]
\centering
\setlength{\tabcolsep}{2pt}
\adjustbox{max width=0.45\textwidth}{%
\begin{tabular}{lcccc}
\toprule
     &
  \begin{tabular}[c]{@{}c@{}}SSIM $ \uparrow$\end{tabular}   &
  \begin{tabular}[c]{@{}c@{}}PSNR-M $\uparrow$\end{tabular} &
  \begin{tabular}[c]{@{}c@{}}LPIPS $ \downarrow$\end{tabular} \\ 

\cmidrule(lr){1-4} 
FOMM\cite{siarohin2019first}   & 0.690 & 19.2  & 0.112 \\
OSFV-S\cite{wang2021facevid2vid}  & 0.807 & \textbf{23.2}  & 0.088 \\
DaGAN\cite{hong2022depth}    & 0.748 & 21.8  & 0.092 \\ \cmidrule(lr){1-4} \rowcolor{Gray}
ROME\cite{Khakhulin2022ROME}     & 0.833  & 21.6 & 0.085 \\
\rowcolor{Gray}
FNeVR\cite{zeng2022fnevr} & 0.801 &  21.1 & 0.092 \\ \cmidrule(lr){1-4} 
Ours     & \textbf{0.862} &  21.9  & \textbf{0.084} \\ 
\bottomrule
\end{tabular}
}
% \vspace{-8pt}
\caption{Comparisons with prior works on self-reenactment (\textbf{quality metrics}) on the VoxCeleb1 dataset\cite{Nagrani17}.
($\uparrow$ indicates larger is better,  while $\downarrow$ indicates smaller is better.) 
}
\label{tab:self_reenactment}
\vspace{-8pt}
\end{table}
\begin{table}[t]
\centering
\setlength{\tabcolsep}{2pt}
\adjustbox{max width=0.4\textwidth}{%
\begin{tabular}{lcccc}
\toprule
 &
  \begin{tabular}[c]{@{}c@{}}CSIM $ \uparrow$\end{tabular} &
  \begin{tabular}[c]{@{}c@{}}AUCON $\uparrow$\end{tabular} &
  \begin{tabular}[c]{@{}c@{}}PRMSE $\downarrow$\end{tabular} &
  \begin{tabular}[c]{@{}c@{}}AVD $ \downarrow$\end{tabular} \\ 

\cmidrule(lr){1-5} 
FOMM\cite{siarohin2019first}   & 0.837 & 0.872 & 2.88 & 0.021 \\
OSFV-S\cite{wang2021facevid2vid}  & 0.911 & 0.934 & 1.81 & 0.014 \\
DaGAN\cite{hong2022depth}    & 0.875 & 0.921  & 1.79 &  0.016 \\ \cmidrule(lr){1-5} \rowcolor{Gray}
ROME\cite{Khakhulin2022ROME}     & 0.906 & 0.918 & 1.68 & 0.013 \\
\rowcolor{Gray}
FNeVR\cite{zeng2022fnevr} & 0.880 &  0.929 & 2.22 & 0.016 \\ \cmidrule(lr){1-5} 
Ours     & \textbf{0.931} &    \textbf{0.956}    &  \textbf{1.66}   &  \textbf{0.010}  \\ 
\bottomrule
\end{tabular}
}
% \vspace{-8pt}
\caption{Comparisons with prior works on self-reenactment (\textbf{fidelity metrics}) on the VoxCeleb1 dataset\cite{Nagrani17}.
% ($\uparrow$ indicates larger is better,  while $\downarrow$ indicates smaller is better.)
}
\vspace{-16pt}
\label{tab:self_reenactment_fidelity}
\end{table}
\begin{figure}[!ht]
  \centering
  \includegraphics[width=\linewidth]{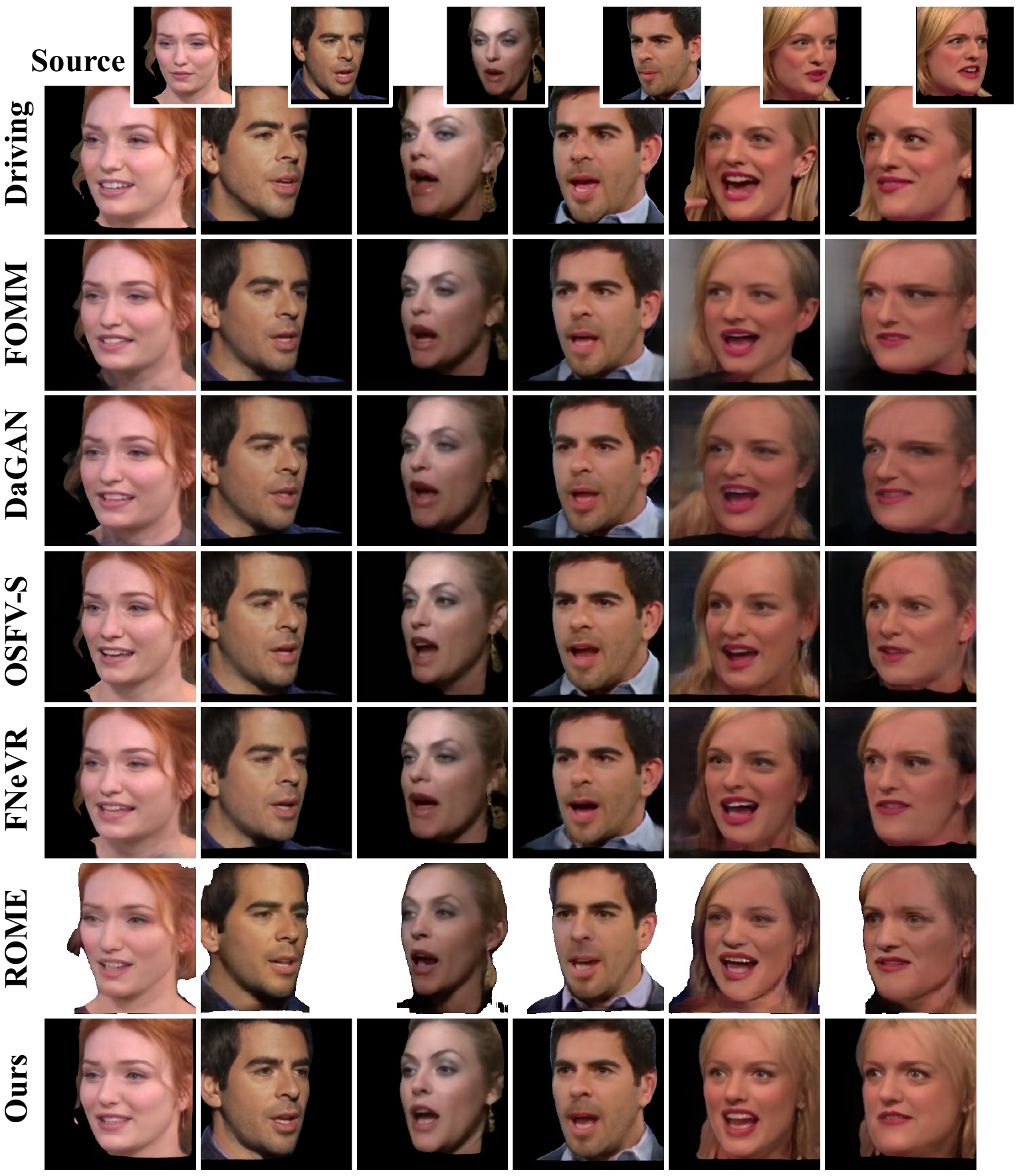}
  % \vspace{-6mm}
  \caption{
\textbf{Qualitative comparisons of self-reenactment on the VoxCeleb1 dataset\cite{Nagrani17}. } 
% Our method can better capture the motion of the driving video and preserve the identity of the source video under large head pose changes. 
}
\vspace*{-16pt}
\label{fig:self_reenactment}
\end{figure}

\begin{table*}[t]
\centering
\setlength{\tabcolsep}{2pt}
\adjustbox{max width=\textwidth}{%

\begin{tabular}{l|ccccc|ccccc|ccccc}
\toprule
 & \multicolumn{5}{c|}{VoxCeleb1~\cite{Nagrani17}} & \multicolumn{5}{c|}{VoxCeleb2~\cite{Chung18b}} & \multicolumn{5}{c}{TalkingHead-1KH~\cite{wang2021facevid2vid}} \\
 &CSIM $\uparrow$ &AUCON $\uparrow$ &PRMSE $\downarrow$  &FID $\downarrow$ &AVD $\downarrow$ &CSIM $\uparrow$ &AUCON $\uparrow$ &PRMSE $\downarrow$ &FID $\downarrow$  &AVD $\downarrow$ &CSIM $\uparrow$ &AUCON $\uparrow$ &PRMSE $\downarrow$ &FID $\downarrow$  &AVD $\downarrow$ \\ 
\cmidrule(lr){1-6} \cmidrule(lr){7-11} \cmidrule(lr){12-16}   
FOMM\cite{siarohin2019first}     
& 0.748 & 0.752  & 3.66 & 86 & 0.044 
& 0.680 & 0.707  & 4.16 & 85 & 0.047 
& 0.723 & 0.741 & 3.71  & 76 & 0.039  \\
OSFV-S\cite{wang2021facevid2vid}   
& 0.791 & 0.893  & 3.01 & 74 & 0.028
& 0.711 & 0.833  & 3.84 & 72 & 0.033 
& 0.787 & 0.884  & 3.03 & 67 & 0.025  \\
DaGAN\cite{hong2022depth}          
& 0.790 & 0.880  & 3.06 & 87 & 0.036 
& 0.693 & 0.815  & 3.93 & 86 & 0.040 
& 0.766 & 0.872  & 2.98 & 73 & 0.035   \\ 
% TPSM\cite{zhao2022tpsm}          
% & -- & 0.886  & 2.99 & 67 & -- 
% & -- & --  & -- & 78 & -- 
% & -- & --  & -- & -- & --   \\ 
\cmidrule(lr){1-6} \cmidrule(lr){7-11} \cmidrule(lr){12-16}   
\rowcolor{Gray}ROME\cite{Khakhulin2022ROME}  
& 0.833  & 0.871 & 2.64  & 76 & 0.016 
& 0.710  & 0.821 & 3.08  & 76 & 0.019 
& 0.781  & 0.864 & 2.66  & 68 & 0.017  \\
\rowcolor{Gray}FNeVR\cite{zeng2022fnevr}    
& 0.812  & 0.884 & 3.32  & 82 & 0.041
& 0.699  & 0.829 & 3.90  & 84 & 0.047 
& 0.775  & 0.879 & 3.39  & 73 & 0.037  \\ 
\cmidrule(lr){1-6} \cmidrule(lr){7-11} \cmidrule(lr){12-16}    
Ours     
& \textbf{0.876} & \textbf{0.917}  & \textbf{2.62}  & \textbf{57} & \textbf{0.012}  
& \textbf{0.787} & \textbf{0.889}  & \textbf{2.91}  & \textbf{61} & \textbf{0.014} 
& \textbf{0.828} & \textbf{0.901}  & \textbf{2.60}  & \textbf{52} & \textbf{0.011}  
\\ 
\bottomrule
\end{tabular}
}
% \vspace{-6pt}
\caption{Comparisons with prior works on cross-identity reenactment with different datasets.
}
\vspace{-12pt}
\label{tab:cross_id_combined}
\end{table*}

\noindent\textbf{Self-Reenactment. } \hspace{-0.3cm}
We first compare the synthesized results when the source and driving images are of the same person. 
The quantitative results concerning the quality and fidelity are listed in Tab.~\ref{tab:self_reenactment} and Tab.~\ref{tab:self_reenactment_fidelity}. 
It can be observed that our method outperforms other state-of-the-arts on all fidelity metrics. 
Fig.~\ref{fig:self_reenactment} displays the qualitative comparisons. 
When head rotation is minimal, all generated results are of similar quality, but as rotation increases, our method produces images with significantly better quality.  \looseness=-1 
% \wang{similar results when pose changes is small, better results under large pose changes}

\begin{figure*}[t]
  \centering
  \includegraphics[width=0.95\linewidth]{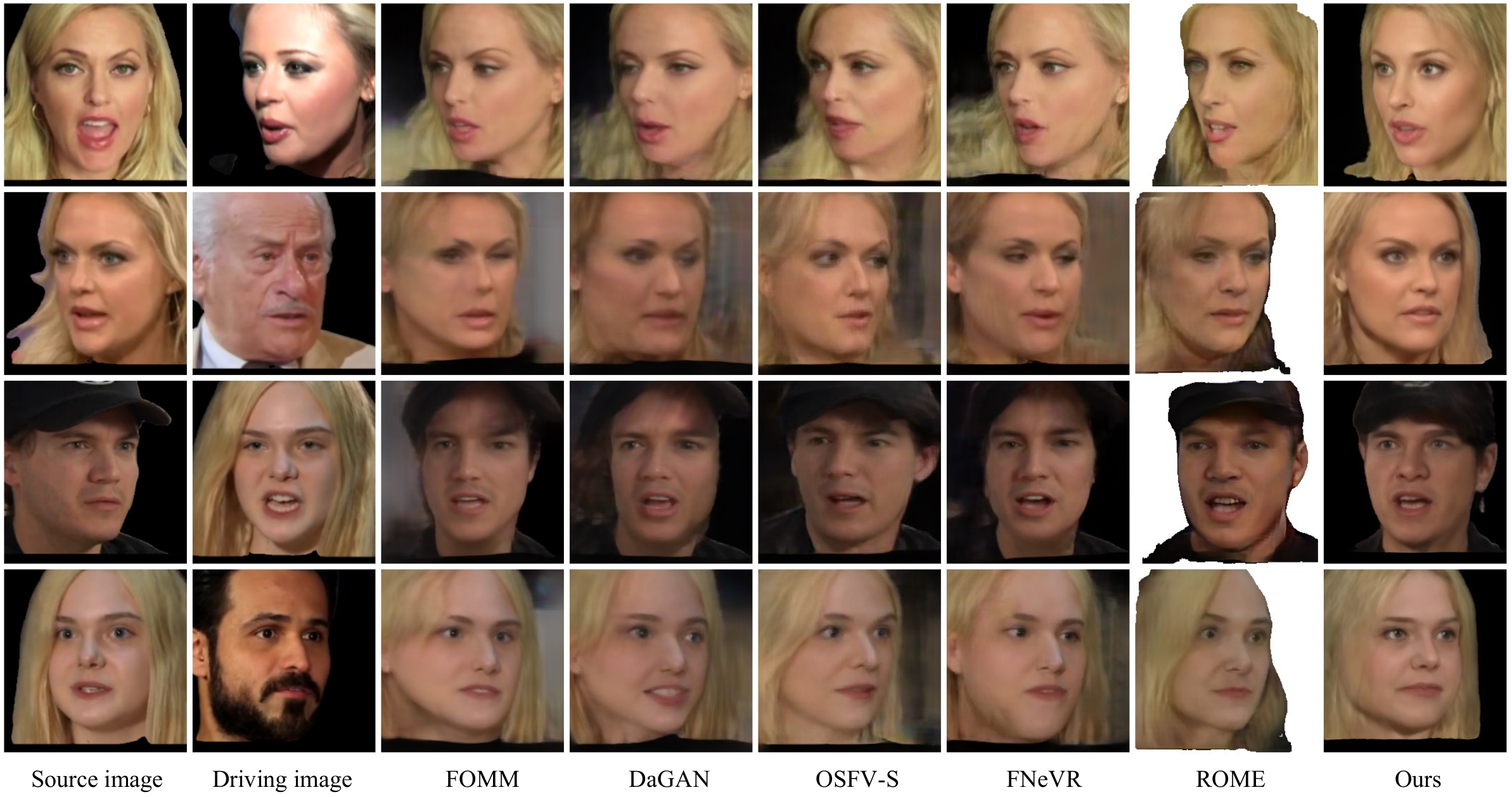}
\vspace{-3pt}
\caption{
\textbf{Qualitative comparisons of cross-identity reenactment on the VoxCeleb1 dataset\cite{Nagrani17}. } Our method captures the driving motion and preserves the identity information better.
}
% \vspace{-12pt}
  \label{fig:qualitative_res}
\end{figure*}

% \vspace{0.1cm}
\noindent\textbf{Cross-Identity Reenactment.} 
\hspace{-1mm}
% \wang{emphasize large pose results comparison}
% We also perform experiments on the test set of VoxCeleb1~\cite{Nagrani17} and TalkingHead-1KH~\cite{wang2021facevid2vid} to exploit the cross-identity motion transfer, where the source image and the driving image contains different persons.
The cross-identity motion transfer was performed on VoxCeleb1~\cite{Nagrani17}, VoxCeleb2~\cite{Chung18b} and TalkingHead-1KH~\cite{wang2021facevid2vid}, where the source image and the driving image contains different persons.
% We also perform experiments on the test set of VoxCeleb1~\cite{Nagrani17} and TalkingHead-1KH~\cite{wang2021facevid2vid} to exploit the cross-identity motion transfer, where the source image and the driving image contains different persons.
% The quantitative results are reported as in Table.~\ref{tab:cross_id_combined}. 
% Our method obtains the best scores in these three metrics \wrt three different datasets, clearly conﬁrming our proposed \textit{MGA} and \textit{LED} benefit the fidelity. 
The quantitative results, as detailed in Tab.~\ref{tab:cross_id_combined}, show that our method outperforms other methods substantially, conclusively affirming the positive impact of our proposed \textit{MGA} and \textit{LED} on image fidelity. 
Furthermore, Fig.~\ref{fig:qualitative_res} presents qualitative results, where the source and driving images differ significantly in several attributes (head orientation, gender, facial shape, skin tone, \etc). 
As shown in the second and third row of Fig.~\ref{fig:qualitative_res}, 
previous works might generate images with artifacts when the source and driving image have large head rotations, while our method can still produce high-fidelity results. 
Furthermore, the generated facial shape of other methods is influenced by the driving image, while our method effectively preserves the source face shape, showing improved identity fidelity. 
Finally, our method exhibits remarkable precision in imitating the expression of the driving image, as demonstrated in the last two rows of Fig.~\ref{fig:qualitative_res}.

\begin{figure}[!ht]
  \centering
  \includegraphics[width=\linewidth]{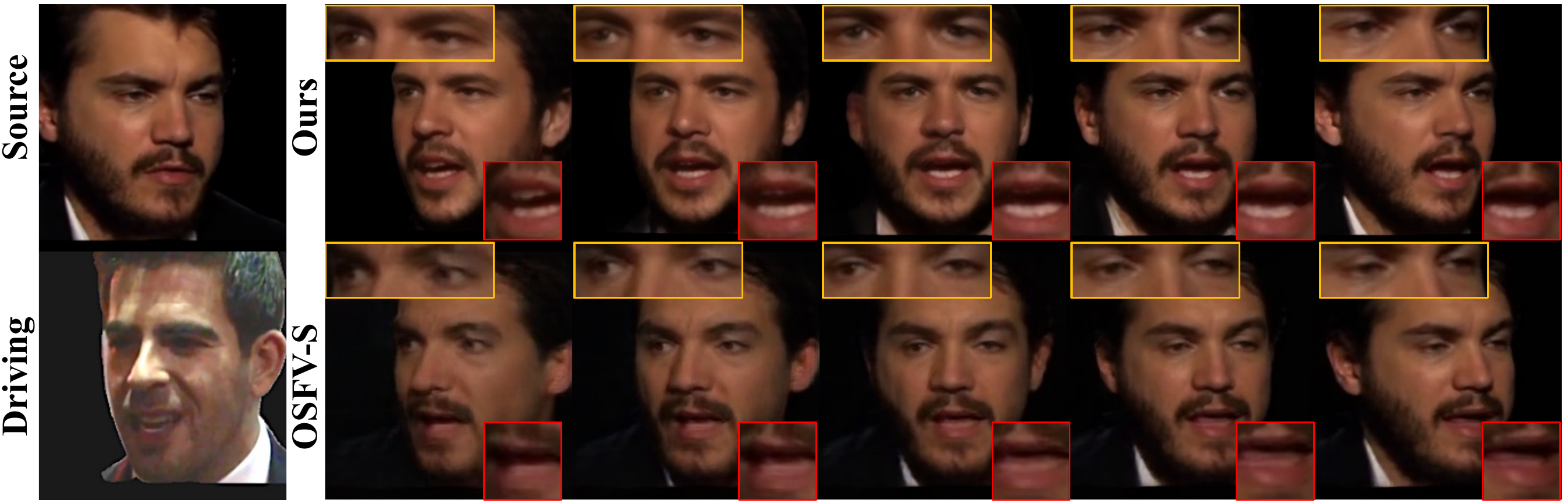}
  % \vspace*{-0.4cm}
  \caption{
\textbf{Qualitative comparisons of free-view generation on the VoxCeleb1 dataset\cite{Nagrani17}. }(Best view when zoomed in).
  }
  \label{fig:free_view}
  % \vspace*{-0.2cm}
\end{figure}
\begin{table}[t]
\centering
\setlength{\tabcolsep}{2pt}
\adjustbox{max width=0.4\textwidth}{%
\begin{tabular}{lccccc}
\toprule
   &
  \begin{tabular}[c]{@{}c@{}}CSIM $\uparrow$\end{tabular}   &
  \begin{tabular}[c]{@{}c@{}}AUCON $\uparrow$\end{tabular}   &
  \begin{tabular}[c]{@{}c@{}}PRMSE $\downarrow$\end{tabular} &
  \begin{tabular}[c]{@{}c@{}}AVD  $\downarrow$\end{tabular} &
  \\ 
\cmidrule(lr){1-5} 
OSFV-S\cite{wang2021facevid2vid}     & 0.727  & 0.808 & 4.82  & 0.039  \\
Ours     & \textbf{0.829} &  \textbf{0.864}  & \textbf{3.78}    & \textbf{0.014} \\ 
\bottomrule
\end{tabular}

}
% \vspace{-8pt}
\caption{Comparisons with OSFV\cite{wang2021facevid2vid} on cross-identity reenactment under free-view generation with the VoxCeleb1 dataset\cite{Nagrani17}.
% ($\uparrow$ indicates larger is better,  while $\downarrow$ indicates smaller is better.)
}
\label{tab:novel_view}
\vspace{-8pt}
\end{table}

% \vspace{-0.1cm}
\subsection{Free-View Synthesis}
% \vspace{-0.1cm}

% \wang{explain the meaning of results number in text, not just listing, they are already listed in tables.}
We also benchmark the face redirection capability of our proposed Hide-NeRF with other free-view talking head synthesis method~\cite{wang2021facevid2vid}. 
%The experiment is based on cross-identity reenactment, and we generate images with different orientations. 
%Specifically, images are generated with view angles of different yaw and pitch.
Specifically, we render the generated results with different view angles.
% among $[-20, -10, 0, 10, 20]$. 
In Tab.~\ref{tab:novel_view}, we evaluate by rendering results from varying views and averaging the corresponding metrics. 
Our method overtakes OSFV-S in free-view synthesis to a greater extent than cross-identity reenactment,  indicating superior face redirection capability compared to OSFV-S.  
The difference in CSIM is the most noticeable, where our method surpasses OSFV-S by 0.102, demonstrating that our method can well preserve identity information under different views. 
% \wang{which demonstrate xxx}.  
% Please note that the PRMSE is calculated with respect to the view angle because no ground truth image containing the corresponding pose is available, thus resulting in a larger PRMSE value.
We also exhibit some qualitative comparisons with OSFV-S on the yaw angle extrapolation in Fig.~\ref{fig:free_view}. 
Our method produces more realistic results.
% As shown in Fig.~\ref{fig:free_view}, the mouth shape of our generated result is more similar to the driving image. 
% Also, our generated images preserve  realistic details such as teeth. 
As seen in Fig.~\ref{fig:free_view}, our method accurately imitates the mouth shape of the driving image and preserves realistic details such as teeth.
Despite large divergence from the source image in the leftmost column, our method maintains skin texture, while OSFV-S fails to do so.
% Besides, our method maintains the expression-consistency under different view angles. 
Our method also retains expression consistency under differing view angles, unlike OSFV-S, which has mismatched eye gaze.
Please consult the supplementary for more qualitative comparisons. \looseness=-1

% \vspace{-0.1cm}
\subsection{Ablation Study}\label{sec:ablation}
% \vspace{-0.1cm}

We also benchmark our performance gain upon different modules. 
Specifically, we conduct four ablations about our proposed \textit{MGA} and \textit{LED}. 
As for the \textit{MGA}, we replace our proposed multi-scale tri-plane representation with single-scale tri-plane representation (w/o multi-scale). 
We also test the effectiveness of the camera to world transformation by deprecating it. 
Concerning the \textit{LED}, we deploy a pose-coupled SECC (w/o SECC), with $\mathbf{R} = \mathbf{R}^{dri}, \mathbf{t} = \mathbf{t}^{dri}$. 

\begin{table}[t]
\centering
\setlength{\tabcolsep}{2pt}
\adjustbox{max width=0.45\textwidth}{%
\begin{tabular}{lcccc}

\toprule
 &
  \begin{tabular}[c]{@{}c@{}}CSIM  $\uparrow$\end{tabular} &
  \begin{tabular}[c]{@{}c@{}}AUCON $\uparrow$\end{tabular} &
  \begin{tabular}[c]{@{}c@{}}PRMSE $\downarrow$\end{tabular} &
  \begin{tabular}[c]{@{}c@{}}AVD   $\downarrow$\end{tabular} 
  \\ 
\cmidrule(lr){1-5} 
w/o multi-scale
& 0.814 &  0.852  & \textbf{2.62} & 0.017 \\
w/o cam2world
& 0.760 &  0.864  &  3.58  & 0.024 \\
w/o SECC  
& 0.802 & 0.879 & 3.06 & 0.018  \\ 
Full     
& \textbf{0.876} &  \textbf{0.917}  &   \textbf{2.62}  & \textbf{0.012} \\ 
\bottomrule
\end{tabular}
}
% \vspace{-8pt}
\caption{Ablation Study over different modules. }
\vspace*{-18pt}
\label{tab:ablation}
\end{table}

\vspace{0.1cm}
\noindent\textbf{Effectiveness of Multi-Scale Tri-Plane Representations.} 
% We show that the tri-plane representation has better expressiveness compared with the single-plane representation, it brought performance gain over all the listed metrics. 
Multi-scale representation brought more delicate features concerning the identity (CSIM increased by 0.062) and expression details (AUCON increased by 0.065), but not significant improvement for the head orientation(PRMSE). 
As discussed, the camera-to-world transformation will  relieve the learning difficulty of the tri-plane representation, thus brought much performance gain regarding different metrics.  

\noindent\textbf{Effectiveness of Decoupling Pose.} 
% \wang{analyze free-view performance}
As shown in Tab.~\ref{tab:ablation}, coupling the pose with other information will be harmful for the identity preservation (CSIM dropped by 0.074, AVD increased by 0.006) and expression preciseness (AUCON dropped by 0.038). 
% The shape preservation is also affected (). 
% The intuition lies in the fact that different poses lead to different shape representations. 
These results verify that our proposed module benefits the fidelity of talking-head generation. 
% Please refer to the Supplementary Materials for more results. 
Please consult the Supplementary Materials for more qualitative comparisons about the ablation studies. 
\section{Conclusion}
\vspace{-4pt}
%\noindent\textbf{Limitations.} 
%\hspace{-1mm}
%Our method has certain limitations. 
%First, like all other NeRF-based methods, the quality of camera registration affects the results. 
%Second, there is no existing dataset containing a sufficient number of head poses so that we can integrate the background into the rendering results. 
%Also, considering the social impact, being a face reenactment method has the risk of misuse for "DeepFakes."  
In this paper, we propose High-fidelity and Deformable Neural Radiance Field (HiDe-NeRF) for high-fidelity and free-view talking head synthesis.
HiDe-NeRF learns a multi-scale neural radiance field from one source image to preserve identity information, and use a expression-aware deformation field to model local non-rigid expression. Ablation studies clearly show that the proposed modules can benefit the motion transfer between two faces.
We demonstrate that our approach can achieve the state-of-the-art synthesis quality on multiple benchmark datasets. 
%Another advantage of our model is that the driver of our deformation module is SECC, which can be obtained from different modalities (image, audio, text, \etc). 
% We consider this virtue as fruitful avenues for future work. 
Moreover, our model can be easily applied to other modality-driven (audio, text, \etc) talking head synthesis, by replacing the inputs of the expression-aware deformation module. 
We consider this virtue as fruitful avenues for future work. 

\vspace{0.1cm}
\noindent\textbf{Limitations.} 
\hspace{-1mm}
Our method has certain limitations. 
First, we can't handle the obvious facial occlusions in the source image. 
Second, due to the pose bias in the training datasets, we can not obtain satisfactory results with extreme poses. 
Also, considering the social impact, being a face reenactment method has the risk of misuse for ``DeepFakes".

\vspace{0.1cm}
\noindent{\textbf{Acknowledgement. }}
This research is supported in part by
Shanghai AI Laboratory, National Natural Science Foundation of China under Grant 62106183 and 62106182, Natural Science Basic Research Program of Shaanxi
under Grant 2021JQ-204, and Alibaba Group.

\clearpage

%%%%%%%%% REFERENCES
{\small
\bibliographystyle{ieee_fullname}
\bibliography{egbib}

\begin{thebibliography}{10}\itemsep=-1pt

\bibitem{athar2022rignerf}
ShahRukh Athar, Zexiang Xu, Kalyan Sunkavalli, Eli Shechtman, and Zhixin Shu.
\newblock Rignerf: Fully controllable neural 3d portraits.
\newblock In {\em IEEE Conf. Comput. Vis. Pattern Recog.}, pages 20364--20373,
  2022.

\bibitem{averbuch2017bringing}
Hadar Averbuch-Elor, Daniel Cohen-Or, Johannes Kopf, and Michael~F Cohen.
\newblock Bringing portraits to life.
\newblock In {\em ACM Trans. Graph.}, 2017.

\bibitem{bergman2022gnarf}
Alexander~W. Bergman, Petr Kellnhofer, Wang Yifan, Eric~R. Chan, David~B.
  Lindell, and Gordon Wetzstein.
\newblock Generative neural articulated radiance fields.
\newblock In {\em Adv. Neural Inform. Process. Syst.}, 2022.

\bibitem{blanz1999morphable}
Volker Blanz and Thomas Vetter.
\newblock A morphable model for the synthesis of 3d faces.
\newblock In {\em Proceedings of the 26th annual conference on Computer
  graphics and interactive techniques}, pages 187--194, 1999.

\bibitem{bulat2017far}
Adrian Bulat and Georgios Tzimiropoulos.
\newblock How far are we from solving the 2d \& 3d face alignment problem? (and
  a dataset of 230,000 3d facial landmarks).
\newblock In {\em Int. Conf. Comput. Vis.}, 2017.

\bibitem{burkov2020neural}
Egor Burkov, Igor Pasechnik, Artur Grigorev, and Victor Lempitsky.
\newblock Neural head reenactment with latent pose descriptors.
\newblock In {\em IEEE Conf. Comput. Vis. Pattern Recog.}, 2020.

\bibitem{Chan2022}
Eric~R. Chan, Connor~Z. Lin, Matthew~A. Chan, Koki Nagano, Boxiao Pan,
  Shalini~De Mello, Orazio Gallo, Leonidas Guibas, Jonathan Tremblay, Sameh
  Khamis, Tero Karras, and Gordon Wetzstein.
\newblock Efficient geometry-aware {3D} generative adversarial networks.
\newblock In {\em IEEE Conf. Comput. Vis. Pattern Recog.}, 2022.

\bibitem{Chung18b}
J.~S. Chung, A. Nagrani, and A. Zisserman.
\newblock Voxceleb2: Deep speaker recognition.
\newblock In {\em INTERSPEECH}, 2018.

\bibitem{deng2019accurate}
Yu Deng, Jiaolong Yang, Sicheng Xu, Dong Chen, Yunde Jia, and Xin Tong.
\newblock Accurate 3d face reconstruction with weakly-supervised learning: From
  single image to image set.
\newblock In {\em IEEE Conf. Comput. Vis. Pattern Recog. Worksh.}, 2019.

\bibitem{dong2018soft}
Haoye Dong, Xiaodan Liang, Ke Gong, Hanjiang Lai, Jia Zhu, and Jian Yin.
\newblock Soft-gated warping-{GAN} for pose-guided person image synthesis.
\newblock In {\em Adv. Neural Inform. Process. Syst.}, 2018.

\bibitem{Drobyshev22MP}
Nikita Drobyshev, Jenya Chelishev, Taras Khakhulin, Aleksei Ivakhnenko, Victor
  Lempitsky, and Egor Zakharov.
\newblock Megaportraits: One-shot megapixel neural head avatars.
\newblock In {\em ACM Int. Conf. Multimedia}, 2022.

\bibitem{geng2018warp}
Jiahao Geng, Tianjia Shao, Youyi Zheng, Yanlin Weng, and Kun Zhou.
\newblock Warp-guided {GANs} for single-photo facial animation.
\newblock In {\em ACM Trans. Graph.}, 2018.

\bibitem{grassal2022neural}
Philip-William Grassal, Malte Prinzler, Titus Leistner, Carsten Rother,
  Matthias Nie{\ss}ner, and Justus Thies.
\newblock Neural head avatars from monocular rgb videos.
\newblock In {\em IEEE Conf. Comput. Vis. Pattern Recog.}, pages 18653--18664,
  2022.

\bibitem{gu2020flnet}
Kuangxiao Gu, Yuqian Zhou, and Thomas~S Huang.
\newblock {FLNet}: Landmark driven fetching and learning network for faithful
  talking facial animation synthesis.
\newblock In {\em AAAI}, 2020.

\bibitem{guo2021adnerf}
Yudong Guo, Keyu Chen, Sen Liang, Yongjin Liu, Hujun Bao, and Juyong Zhang.
\newblock Ad-nerf: Audio driven neural radiance fields for talking head
  synthesis.
\newblock In {\em Int. Conf. Comput. Vis.}, 2021.

\bibitem{ha2020marionette}
Sungjoo Ha, Martin Kersner, Beomsu Kim, Seokjun Seo, and Dongyoung Kim.
\newblock {MarioNETte}: Few-shot face reenactment preserving identity of unseen
  targets.
\newblock In {\em AAAI}, 2020.

\bibitem{hong2022depth}
Fa-Ting Hong, Longhao Zhang, Li Shen, and Dan Xu.
\newblock Depth-aware generative adversarial network for talking head video
  generation.
\newblock In {\em IEEE Conf. Comput. Vis. Pattern Recog.}, pages 3397--3406,
  2022.

\bibitem{DeepFaceVideoEditing2022}
Kaiwen Jiang, Shu-Yu Chen, Feng-Lin Liu, Hongbo Fu, and Lin Gao.
\newblock {NeRFFaceEditing}: Disentangled face editing in neural radiance
  fields.
\newblock {\em SIGGRAPH Asia}, 2022.

\bibitem{Karras2020ada}
Tero Karras, Miika Aittala, Janne Hellsten, Samuli Laine, Jaakko Lehtinen, and
  Timo Aila.
\newblock Training generative adversarial networks with limited data.
\newblock In {\em Adv. Neural Inform. Process. Syst.}, 2020.

\bibitem{karras2019stylegan}
Tero Karras, Samuli Laine, and Timo Aila.
\newblock A style-based generator architecture for generative adversarial
  networks.
\newblock In {\em IEEE Conf. Comput. Vis. Pattern Recog.}, pages 4401--4410,
  2019.

\bibitem{karras2020stylegan2}
Tero Karras, Samuli Laine, Miika Aittala, Janne Hellsten, Jaakko Lehtinen, and
  Timo Aila.
\newblock Analyzing and improving the image quality of stylegan.
\newblock In {\em IEEE Conf. Comput. Vis. Pattern Recog.}, pages 8110--8119,
  2020.

\bibitem{Khakhulin2022ROME}
Taras Khakhulin, Vanessa Sklyarova, Victor Lempitsky, and Egor Zakharov.
\newblock Realistic one-shot mesh-based head avatars.
\newblock In {\em Eur. Conf. Comput. Vis.}, 2022.

\bibitem{liu2019liquid}
Wen Liu, Zhixin Piao, Jie Min, Wenhan Luo, Lin Ma, and Shenghua Gao.
\newblock {Liquid Warping GAN}: A unified framework for human motion imitation,
  appearance transfer and novel view synthesis.
\newblock In {\em Int. Conf. Comput. Vis.}, 2019.

\bibitem{liu2022semantic}
Xian Liu, Yinghao Xu, Qianyi Wu, Hang Zhou, Wayne Wu, and Bolei Zhou.
\newblock Semantic-aware implicit neural audio-driven video portrait
  generation.
\newblock In {\em Eur. Conf. Comput. Vis.}, 2022.

\bibitem{mallya2022implicit}
Arun Mallya, Ting-Chun Wang, and Ming-Yu Liu.
\newblock {Implicit Warping for Animation with Image Sets}.
\newblock In {\em Adv. Neural Inform. Process. Syst.}, 2022.

\bibitem{mildenhall2020nerf}
Ben Mildenhall, Pratul~P. Srinivasan, Matthew Tancik, Jonathan~T. Barron, Ravi
  Ramamoorthi, and Ren Ng.
\newblock {NeRF}: Representing scenes as neural radiance fields for view
  synthesis.
\newblock In {\em Eur. Conf. Comput. Vis.}, 2020.

\bibitem{Nagrani17}
A. Nagrani, J.~S. Chung, and A. Zisserman.
\newblock Voxceleb: a large-scale speaker identification dataset.
\newblock In {\em INTERSPEECH}, 2017.

\bibitem{park2021nerfies}
Keunhong Park, Utkarsh Sinha, Jonathan~T Barron, Sofien Bouaziz, Dan~B Goldman,
  Steven~M Seitz, and Ricardo Martin-Brualla.
\newblock Nerfies: Deformable neural radiance fields.
\newblock In {\em Int. Conf. Comput. Vis.}, pages 5865--5874, 2021.

\bibitem{park2021hypernerf}
Keunhong Park, Utkarsh Sinha, Peter Hedman, Jonathan~T. Barron, Sofien Bouaziz,
  Dan~B Goldman, Ricardo Martin-Brualla, and Steven~M. Seitz.
\newblock {HyperNeRF}: A higher-dimensional representation for topologically
  varying neural radiance fields.
\newblock In {\em ACM Trans. Graph.} ACM, dec 2021.

\bibitem{park2019semantic}
Taesung Park, Ming-Yu Liu, Ting-Chun Wang, and Jun-Yan Zhu.
\newblock Semantic image synthesis with spatially-adaptive normalization.
\newblock In {\em IEEE Conf. Comput. Vis. Pattern Recog.}, pages 2337--2346,
  2019.

\bibitem{bfm}
Pascal Paysan, Reinhard Knothe, Brian Amberg, Sami Romdhani, and Thomas Vetter.
\newblock A 3d face model for pose and illumination invariant face recognition.
\newblock In {\em 2009 sixth IEEE international conference on advanced video
  and signal based surveillance}, pages 296--301. Ieee, 2009.

\bibitem{peng2020convolutional}
Songyou Peng, Michael Niemeyer, Lars Mescheder, Marc Pollefeys, and Andreas
  Geiger.
\newblock Convolutional occupancy networks.
\newblock In {\em Eur. Conf. Comput. Vis.}, pages 523--540. Springer, 2020.

\bibitem{pumarola2018unsupervised}
Albert Pumarola, Antonio Agudo, Alberto Sanfeliu, and Francesc Moreno-Noguer.
\newblock Unsupervised person image synthesis in arbitrary poses.
\newblock In {\em IEEE Conf. Comput. Vis. Pattern Recog.}, 2018.

\bibitem{ren2021pirenderer}
Yurui Ren, Ge Li, Yuanqi Chen, Thomas~H. Li, and Shan Liu.
\newblock Pirenderer: Controllable portrait image generation via semantic
  neural rendering.
\newblock In {\em Int. Conf. Comput. Vis.}, 2021.

\bibitem{shen2022dfrf}
Shuai Shen, Wanhua Li, Zheng Zhu, Yueqi Duan, Jie Zhou, and Jiwen Lu.
\newblock Learning dynamic facial radiance fields for few-shot talking head
  synthesis.
\newblock In {\em Eur. Conf. Comput. Vis.}, 2022.

\bibitem{siarohin2019first}
Aliaksandr Siarohin, St{\'e}phane Lathuili{\`e}re, Sergey Tulyakov, Elisa
  Ricci, and Nicu Sebe.
\newblock First order motion model for image animation.
\newblock {\em Adv. Neural Inform. Process. Syst.}, 32, 2019.

\bibitem{Siarohin2019monkey}
Aliaksandr Siarohin, Stéphane Lathuilière, Sergey Tulyakov, Elisa Ricci, and
  Nicu Sebe.
\newblock Animating arbitrary objects via deep motion transfer.
\newblock In {\em IEEE Conf. Comput. Vis. Pattern Recog.}, June 2019.

\bibitem{siarohin2021motion}
Aliaksandr Siarohin, Oliver Woodford, Jian Ren, Menglei Chai, and Sergey
  Tulyakov.
\newblock Motion representations for articulated animation.
\newblock In {\em IEEE Conf. Comput. Vis. Pattern Recog.}, 2021.

\bibitem{suwajanakorn2017synthesizing}
Supasorn Suwajanakorn, Steven~M Seitz, and Ira Kemelmacher-Shlizerman.
\newblock Synthesizing {Obama}: learning lip sync from audio.
\newblock In {\em ACM Trans. Graph.}, 2017.

\bibitem{tewari2020stylerig}
Ayush Tewari, Mohamed Elgharib, Gaurav Bharaj, Florian Bernard, Hans-Peter
  Seidel, Patrick P{\'e}rez, Michael Z{\"o}llhofer, and Christian Theobalt.
\newblock {StyleRig}: Rigging stylegan for 3d control over portrait images.
\newblock In {\em IEEE Conf. Comput. Vis. Pattern Recog.}, 2020.

\bibitem{tewari2020pie}
Ayush Tewari, Mohamed Elgharib, Mallikarjun~B R, Florian Bernard, Hans-Peter
  Seidel, Patrick P\'{e}rez, Michael Zollh\"{o}fer, and Christian Theobalt.
\newblock Pie: Portrait image embedding for semantic control.
\newblock In {\em ACM Trans. Graph.}, 2020.

\bibitem{thies2019deferred}
Justus Thies, Michael Zollh{\"o}fer, and Matthias Nie{\ss}ner.
\newblock Deferred neural rendering: Image synthesis using neural textures.
\newblock In {\em ACM Trans. Graph.}, 2019.

\bibitem{thies2015real}
Justus Thies, Michael Zollh{\"o}fer, Matthias Nie{\ss}ner, Levi Valgaerts, Marc
  Stamminger, and Christian Theobalt.
\newblock Real-time expression transfer for facial reenactment.
\newblock In {\em ACM Trans. Graph.}, 2015.

\bibitem{thies2016face2face}
Justus Thies, Michael Zollhofer, Marc Stamminger, Christian Theobalt, and
  Matthias Nie{\ss}ner.
\newblock {Face2Face}: Real-time face capture and reenactment of rgb videos.
\newblock In {\em IEEE Conf. Comput. Vis. Pattern Recog.}, 2016.

\bibitem{wang2021facevid2vid}
Ting-Chun Wang, Arun Mallya, and Ming-Yu Liu.
\newblock One-shot free-view neural talking-head synthesis for video
  conferencing.
\newblock In {\em IEEE Conf. Comput. Vis. Pattern Recog.}, 2021.

\bibitem{yao2022dfa}
Shunyu Yao, RuiZhe Zhong, Yichao Yan, Guangtao Zhai, and Xiaokang Yang.
\newblock {DFA-NeRF}: Personalized talking head generation via disentangled
  face attributes neural rendering.
\newblock {\em arXiv preprint arXiv:2201.00791}, 2022.

\bibitem{Zakharov20BiLayer}
Egor Zakharov, Aleksei Ivakhnenko, Aliaksandra Shysheya, and Victor Lempitsky.
\newblock Fast bi-layer neural synthesis of one-shot realistic head avatars.
\newblock In {\em Eur. Conf. Comput. Vis.}, August 2020.

\bibitem{zeng2022fnevr}
Bohan Zeng, Boyu Liu, Hong Li, Xuhui Liu, Jianzhuang Liu, Dapeng Chen, Wei
  Peng, and Baochang Zhang.
\newblock {FNeVR}: Neural volume rendering for face animation.
\newblock In {\em Adv. Neural Inform. Process. Syst.}, 2022.

\bibitem{zhang2018unreasonable}
Richard Zhang, Phillip Isola, Alexei~A Efros, Eli Shechtman, and Oliver Wang.
\newblock The unreasonable effectiveness of deep features as a perceptual
  metric.
\newblock In {\em IEEE Conf. Comput. Vis. Pattern Recog.}, pages 586--595,
  2018.

\bibitem{zhao2022tpsm}
Jian Zhao and Hui Zhang.
\newblock Thin-plate spline motion model for image animation.
\newblock In {\em IEEE Conf. Comput. Vis. Pattern Recog.}, pages 3657--3666,
  2022.

\bibitem{zhu2017face}
Xiangyu Zhu, Xiaoming Liu, Zhen Lei, and Stan~Z Li.
\newblock Face alignment in full pose range: A 3d total solution.
\newblock {\em IEEE Trans. Pattern Anal. Mach. Intell.}, 2017.

\bibitem{zhu2022nice}
Zihan Zhu, Songyou Peng, Viktor Larsson, Weiwei Xu, Hujun Bao, Zhaopeng Cui,
  Martin~R Oswald, and Marc Pollefeys.
\newblock Nice-slam: Neural implicit scalable encoding for slam.
\newblock In {\em IEEE Conf. Comput. Vis. Pattern Recog.}, pages 12786--12796,
  2022.

\bibitem{face-parsing}
zllrunning.
\newblock face-parsing.pytorch.
\newblock \url{https://github.com/zllrunning/face-parsing.PyTorch}, 2019.

\end{thebibliography}
}

%%%%%%%%%% Merge with supplemental materials %%%%%%%%%%
\pagebreak
% \widetext
\begin{center}
\textbf{\large Supplemental Material\\One-Shot High-Fidelity Talking-Head Synthesis with Deformable Neural Radiance Field}
\end{center}

%%%%%%%%%% Merge with supplemental materials %%%%%%%%%%
%%%%%%%%%% Prefix a "S" to all equations, figures, tables and reset the counter %%%%%%%%%%
\setcounter{equation}{0}
\setcounter{figure}{0}
\setcounter{table}{0}
\setcounter{page}{1}
\makeatletter
\renewcommand{\theequation}{S\arabic{equation}}
\renewcommand{\thefigure}{S\arabic{figure}}
% \renewcommand{\bibnumfmt}[1]{[S#1]}
% \renewcommand{\citenumfont}[1]{S#1}
%%%%%%%%%% Prefix a "S" to all equations, figures, tables and reset the counter %%%%%%%%%%

In this supplement, we ﬁrst provide additional training details and network architecture implementation for completeness and reproducibility. 
We discuss experiment details, such as datasets and metrics, and provide more qualitative results. 
Lastly, we also show some failure cases that may be targets of future work.

\section{Additional Details about Networks and Training Details}

\begin{figure*}[t]
  \centering
  \includegraphics[width=.8\linewidth]{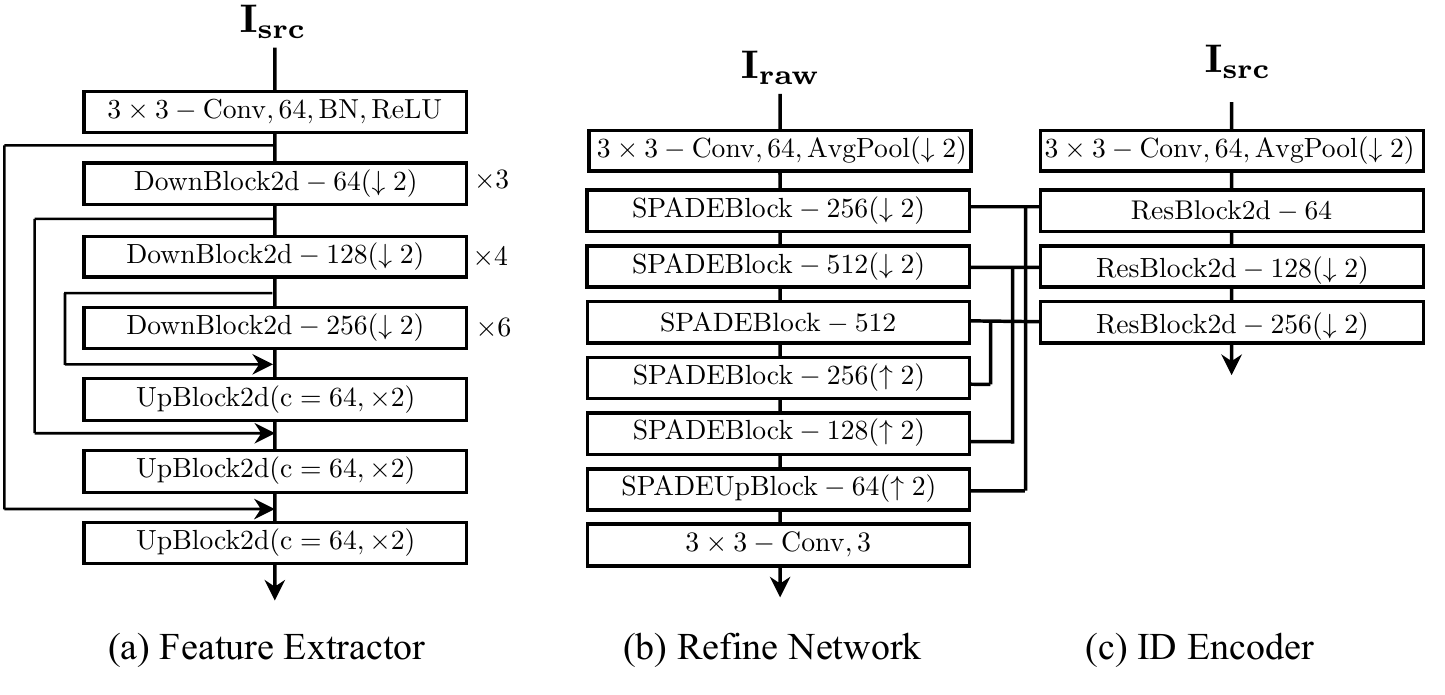}

\vspace{-4mm}
\caption{
\textbf{Illustration of sub-networks.}
  }
  \label{fig:sup_architecture}
  \vspace{-4mm}
\end{figure*}
\begin{figure}[t]
  \centering
  \includegraphics[width=\linewidth]{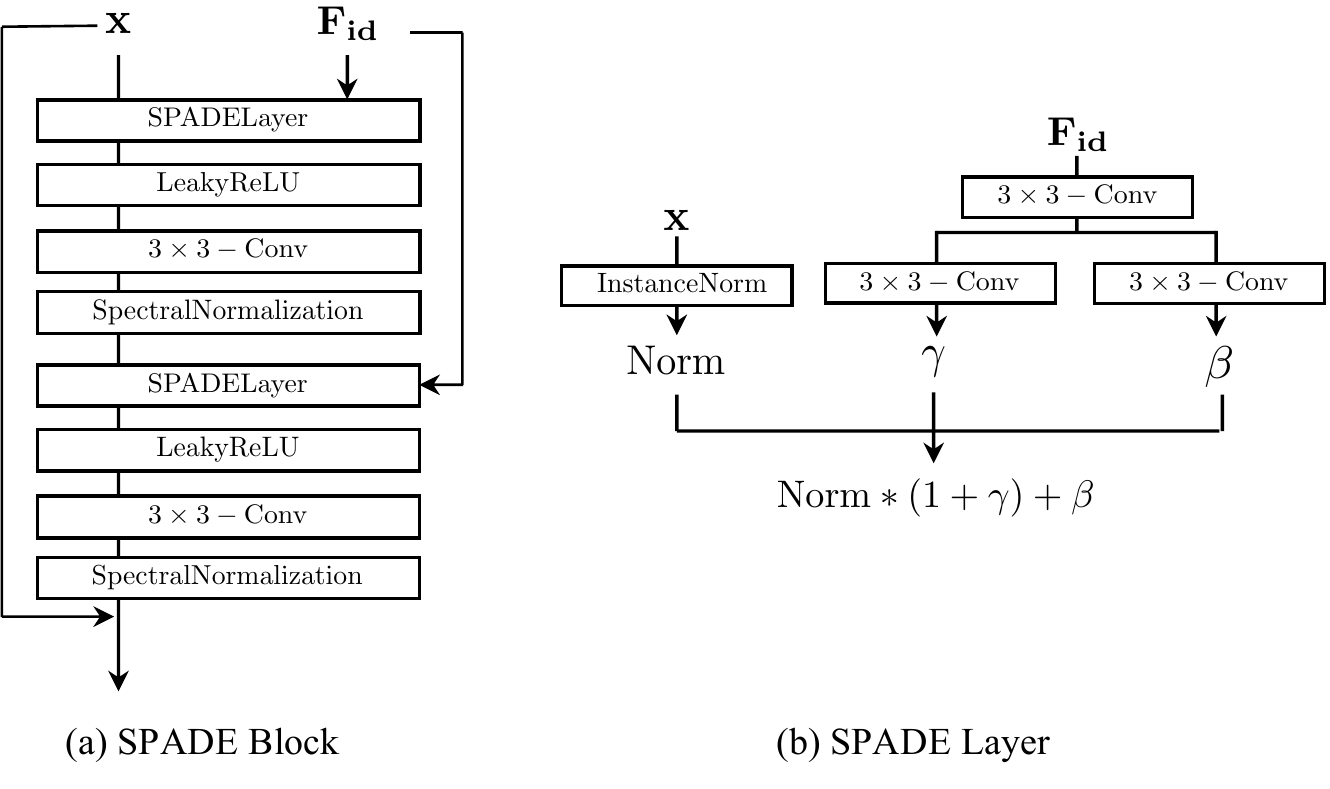}

\vspace{-4mm}
\caption{
\textbf{Illustration of SPADE block and layer.}
  }
  \label{fig:sup_spade}
  \vspace{-4mm}
\end{figure}
\begin{figure}[t]
  \centering
  \includegraphics[width=\linewidth]{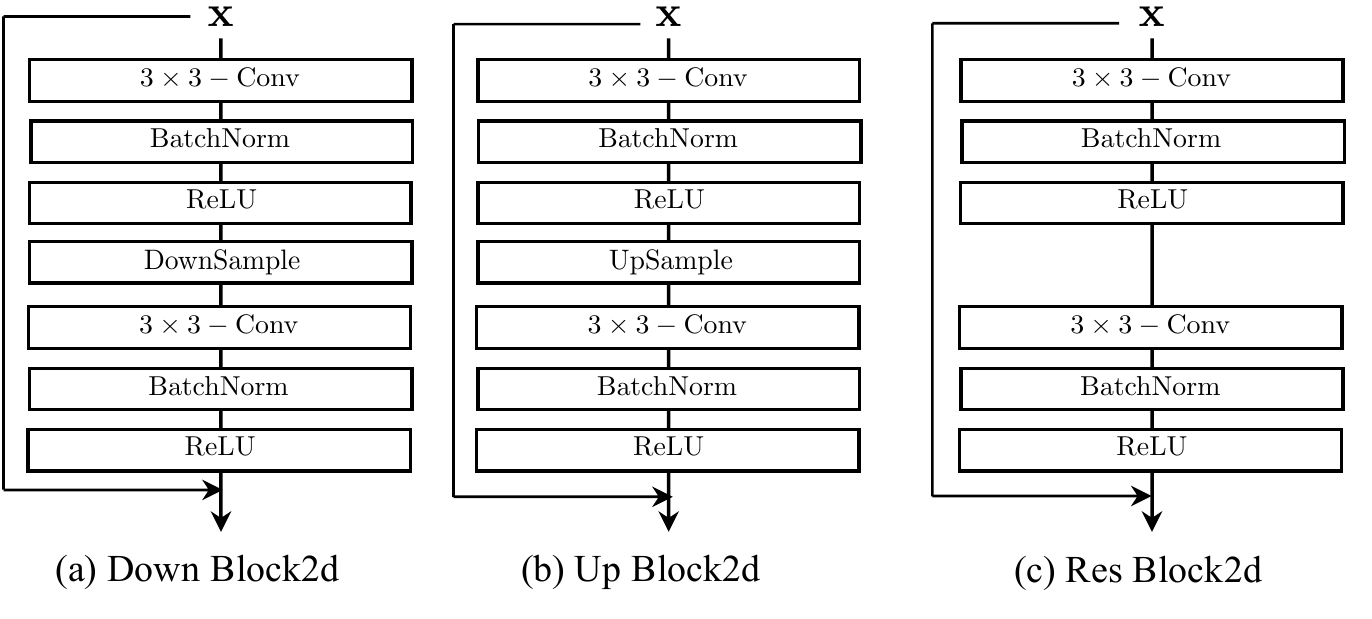}

\vspace{-4mm}
\caption{
\textbf{Illustration of different building blocks.}
  }
  \label{fig:sup_blk}
  \vspace{-4mm}
\end{figure}

\subsection{Training Details}\label{sec:detail}

\noindent\textbf{Perturbation Regularization }
We observe that directly feeding the point positions into the deformation field leads to over-fitting and alias. 
To this end, we devise a perturbation regularization, which adds very small random perturbation into the point positions $\mathbf{P}$ fed into the \textit{LED}. However, the predicted deltas $\Delta \mathbf{P}$ are added to the original points instead of the perturbed points to form the $\mathbf{P}^{\prime}$. 
The intuition of our design lies in the fact that nearby points should have similar deformation, and this perturbation regularization enables our deformation module to learn the \textit{``local consistency"}.

\noindent\textbf{Data Augmentation. }
NeRF is known to benefit from different view inputs, but there are subtle pose changes in most short video clips. 
% Also, subtle pose changes might lead to an over-easy problem during training. 
To improve the diversity of head poses, we set the flipping probability $p_{flip}=0.5$ to randomly flip the source image horizontally. 
Considering flipping leads to changes in the background,
we only preserve the head and torso parts in the video frames via an off-the-shelf segmentation predictor~\cite{face-parsing}.

% \subsection{Loss Details}
\subsection{Loss Functions}\label{sec:loss}

In the training stage, we can take different frames of the same video as the source and driving images, as they share the same identity. 
Our proposed HiDe-NeRF is combined of NeRF and a refine module, thus it will generate two outputs $\mathbf{I_{raw}}$ and $\mathbf{I_{rf}}$, we take the driving image $\mathbf{I_{dri}}$ and its downsampled result $\mathbf{I_{dri}^{down}}$ as their corresponding ground-truth. it is optimized with the following loss functions: 

\noindent\textbf{Mean Square Error Loss $\mathcal{L}_{M}$}.  
We minimize the mean square error of the refined image $\mathbf{I_{rf}}$ and rendered image $\mathbf{I_{raw}}$ \wrt their corresponding ground-truth.

\noindent\textbf{Perceptual Loss $\mathcal{L}_{P}$}. 
We minimize the perceptual loss of the refined image $\mathbf{I_{rf}}$ and rendered image $\mathbf{I_{raw}}$ \wrt their corresponding ground-truth to get a more realistic result.

\noindent\textbf{Adversarial Loss $\mathcal{L}_{G}$}. 
We deploy conditional discriminator as in StyleGAN2-ADA~\cite{Karras2020ada}. 
Specifically, $\mathbf{I_{rf}}$ and $\mathbf{I_{raw}}$ are fed into two separate discriminators, and the camera parameters are used as the condition.

\noindent\textbf{Deformation Regularization $\mathcal{R}_{D}$}. 
We add a deformation regularization as a loss to enforce the deformation module find the ``shortest" path for each single point. 
The deformation regularization is the sum of $l1$ norm for the predicted point-wise delta. 
Concretely, it is calculated by $\mathcal{R}_{D} = \| \mathcal{F}_{\Phi}^{\text {deform }}\left(\mathbf{P}, \mathbf{SECC}_{dri}, \mathbf{SECC}_{can} \right)\|_{1}.$

MSE loss ensures the rendered and refined images be similar to their ground-truth. 
While the perceptual and adversarial loss ensures the images are more realistic. 
And the last regularization loss makes the training stage more stable. 
The overall loss can be summarized as below:

\begin{equation}
\begin{aligned}
\mathcal{L}=& \lambda_M \left(\mathcal{L}_M\left(\mathbf{I}_{rf}, \mathbf{I}_{dri}\right)  + \mathcal{L}_M\left(\{\mathbf{I}_{raw}, \mathbf{I}_{dri}^{down}\}_{n=1}^{K}\right) \right)\\
&+\lambda_P \left(\mathcal{L}_P\left(\mathbf{I}_{rf}, \mathbf{I}_{dri}\right) + \mathcal{L}_P\left(\{\mathbf{I}_{rf}, \mathbf{I}_{dri}^{down}\}_{n=1}^{K}\right) \right) \\
&+\lambda_G \left(\mathcal{L}_G\left(\mathbf{I}_{rf}, \mathbf{I}_{dri}\right) + \mathcal{L}_G\left(\{\mathbf{I}_{rf}, \mathbf{I}_{dri}^{down}\}_{n=1}^{K}\right) \right) \\
&+\lambda_R\mathcal{R}_{D}.
\end{aligned}
\end{equation}

\subsection{Network Architecture details}

The implementation details of some of the sub-networks in our work are illustrated in Fig.~\ref{fig:sup_architecture} and we will introduce them accordingly. 
We also show the architecture of SPADE and different blocks in Fig.~\ref{fig:sup_spade} and Fig.~\ref{fig:sup_blk}. 

\noindent\textbf{Feature Extractor. }
The feature extractor neural network extracts appearance features from the source image. 
It consists of a number of downsampling blocks and a number of upsampling blocks to compute the ﬁnal tri-plane volume features.

\noindent\textbf{Refine Network. } \hspace{-3mm}
In the training process, we feed the rendered image and features extracted by ID Encoder into the refine network, and get the refined image.
The detailed structure of the refine network is shown in Fig.~\ref{fig:sup_architecture}(b). 

\noindent\textbf{ID Encoder. } 
We utilize another convolutional neural network to extract identity information from the source image, and inject the extracted information into the refine network with SPADEBlock. 
The detailed connections between the refine network and the ID encoder is shown in Fig.~\ref{fig:sup_architecture}(b-c).

\section{Additional Details about Experiments}

\subsection{Dataset Details}
We use the following datasets in our evaluations. 
During testing, we first align~\cite{deng2019accurate} faces and segment the facial parts out as the input. 

\vspace{0.1cm}
\noindent\textbf{VoxCeleb1}~\cite{Nagrani17}. 
The VoxCeleb dataset contains about 20,000 videos. 
For pre-processing, we extract an initial bounding box in the ﬁrst video frame. 
The validation dataset contains about 500 videos.

\vspace{0.1cm}
\noindent\textbf{VoxCeleb2}~\cite{Chung18b}. 
This dataset contains about 1M talking head videos of different celebrities. 
We follow the training and test split proposed in the original paper and report our results on the validation set, which contains about 36k videos.

\vspace{0.1cm}
\noindent\textbf{TalkingHead-1KH}~\cite{wang2021facevid2vid}.
This dataset is composed of about 1000 hours of videos from various sources. 
This dataset is in general with higher resolutions and better image quality than those in the VoxCeleb2. 
For fair comparison, we report results on the resized faces at the $256 \times 256$ resolution.

% \subsection{Baselines}
% \subsection{Metrics}

\begin{figure*}[ht]
  \centering
  \includegraphics[width=0.9\linewidth]{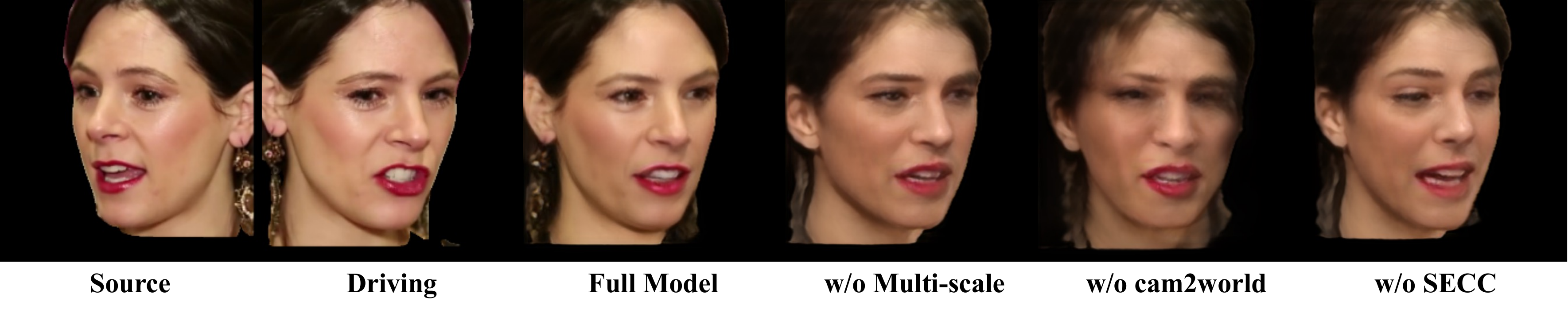}

\vspace{-4mm}
\caption{
\textbf{Ablation Study.}  
  }
  \label{fig:sup_ablation}
  \vspace{-6mm}
\end{figure*}

\vspace{-2mm}
\begin{figure}[b]
  \centering
  \includegraphics[width=0.9\linewidth]{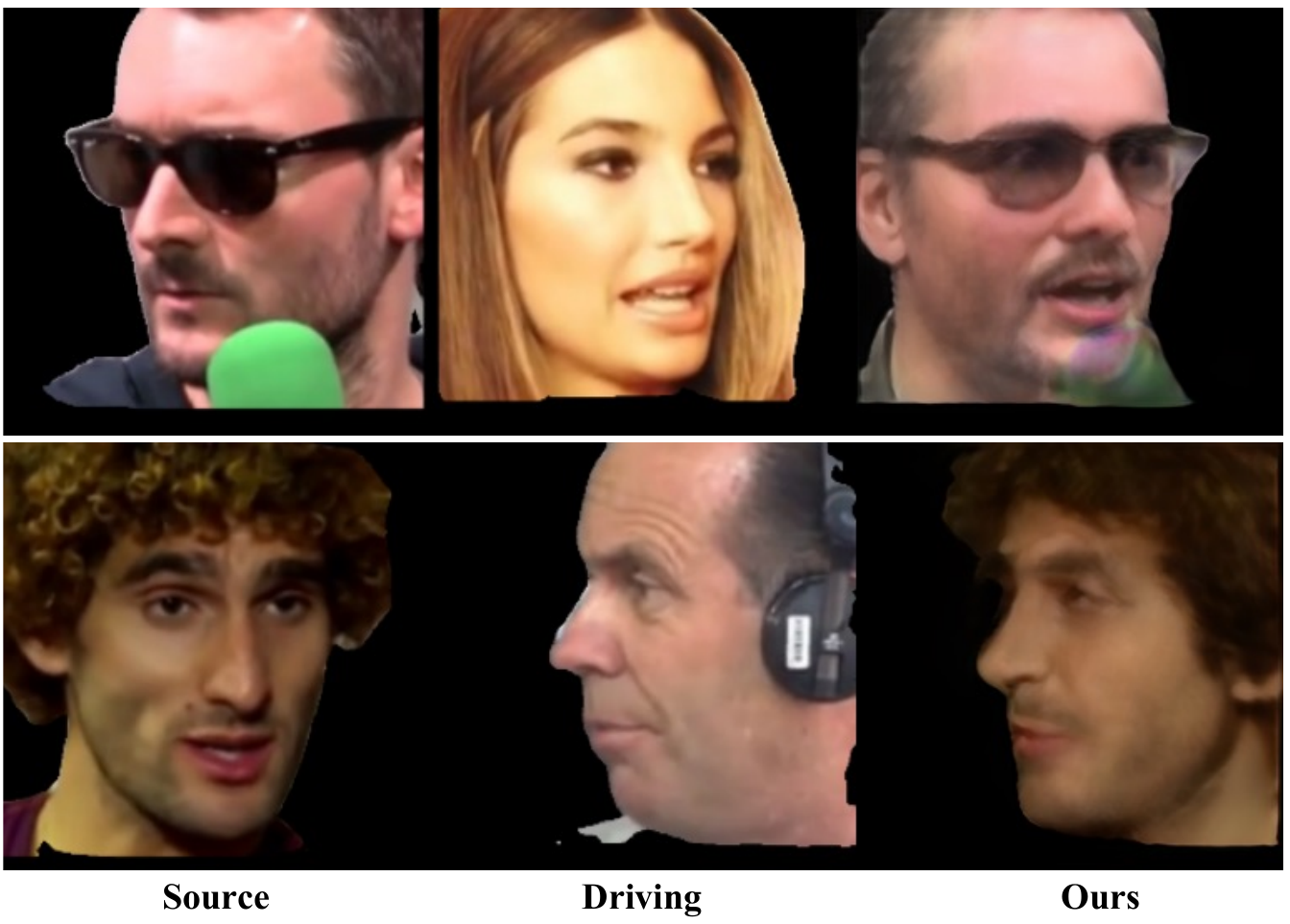}

\vspace{-3mm}
\caption{
\textbf{Example failure cases.} 
  }
  \label{fig:bad}
  \vspace{-4mm}
\end{figure}

\begin{figure*}[ht]
  \centering
  \includegraphics[width=\linewidth]{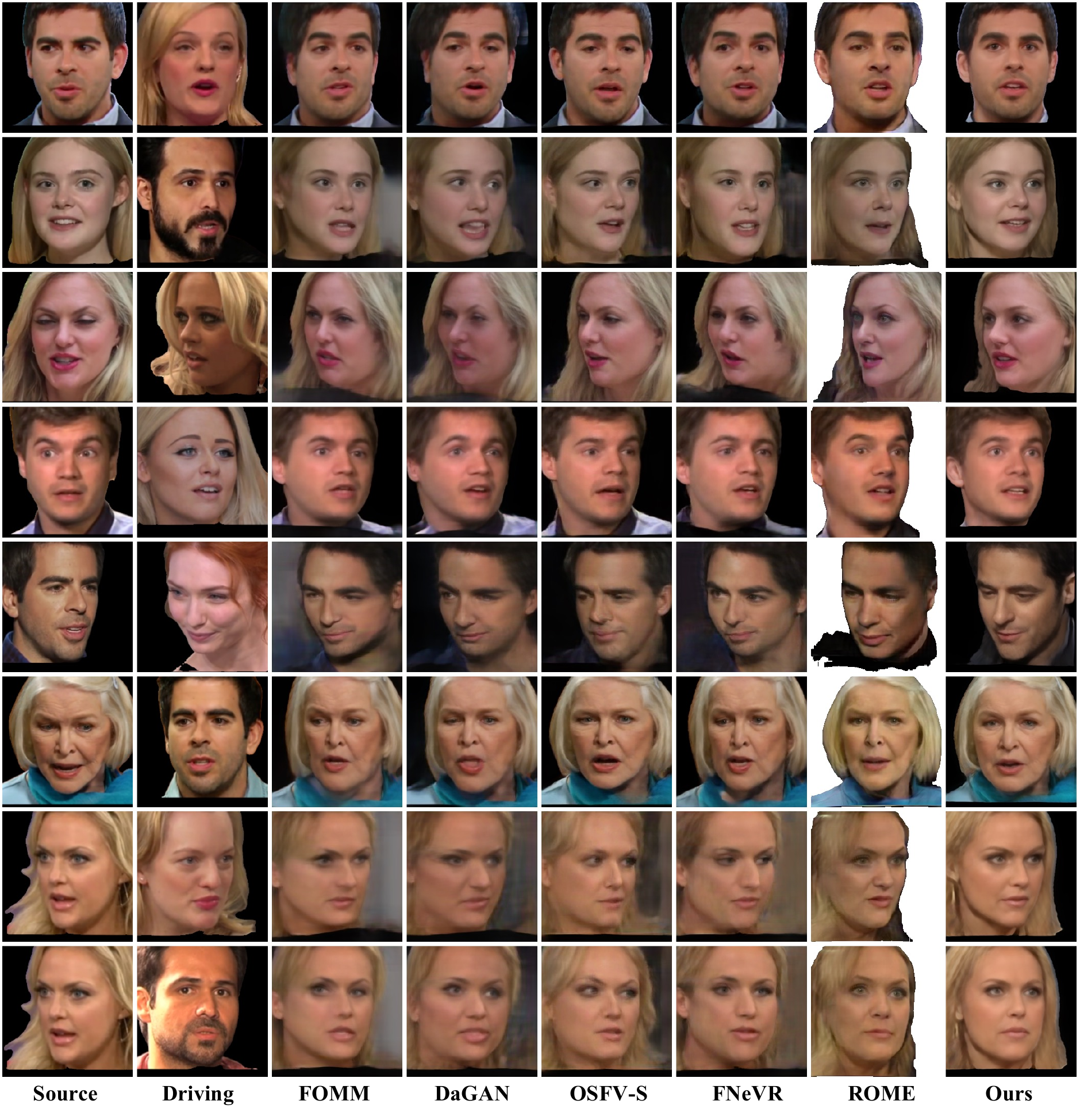}

\caption{
\textbf{Qualitative comparisons of cross-identity reenactment on the VoxCeleb1 dataset\cite{Nagrani17}. }
  }
  \label{fig:sup_1}
  \vspace{-4mm}
\end{figure*}
\begin{figure*}[ht]
  \centering
  \includegraphics[width=\linewidth]{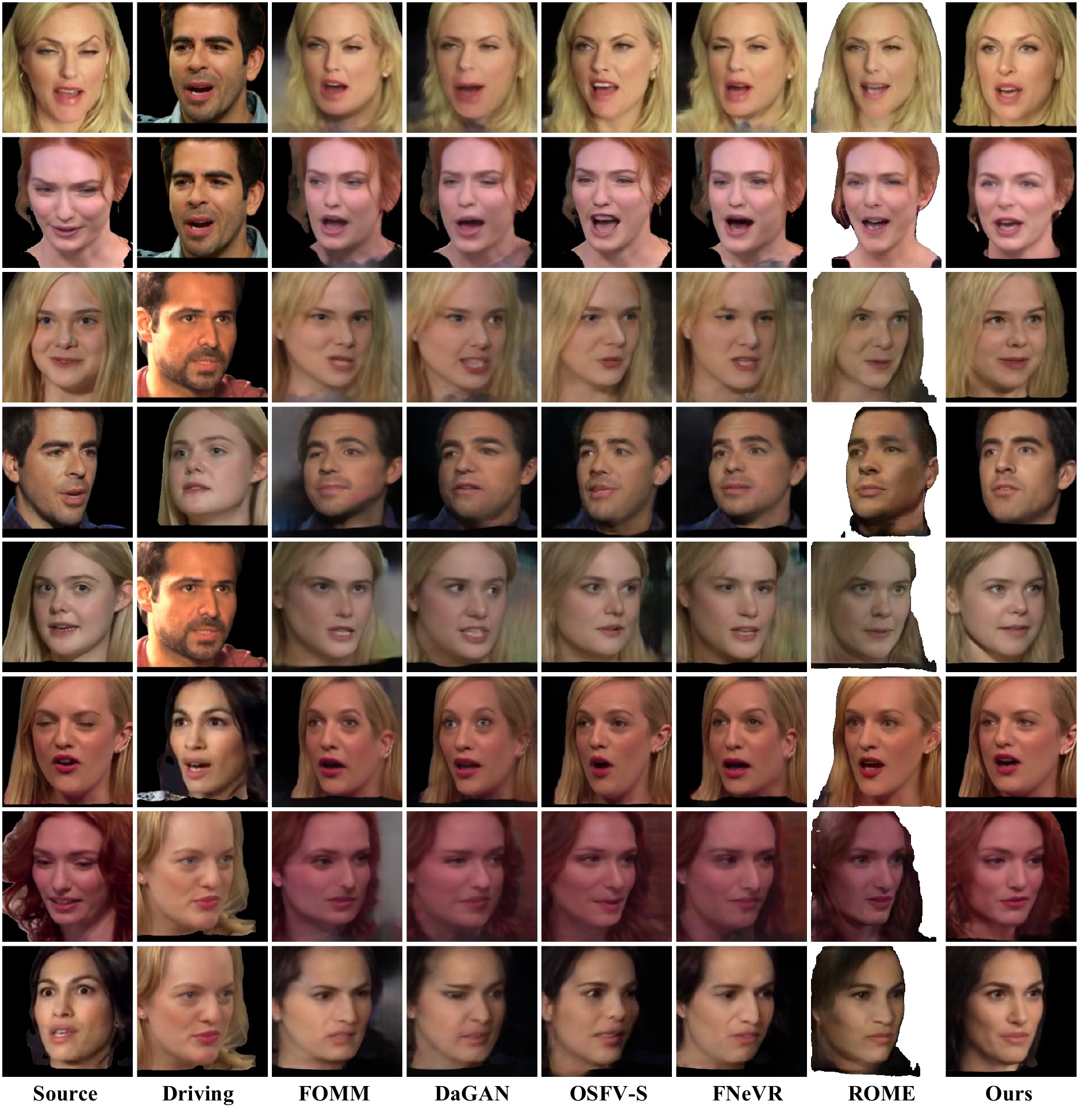}

\caption{
\textbf{Qualitative comparisons of cross-identity reenactment on the VoxCeleb1 dataset\cite{Nagrani17}. }
  }
  \label{fig:sup_2}
  \vspace{-4mm}
\end{figure*}
\begin{figure*}[ht]
  \centering
  \includegraphics[width=\linewidth]{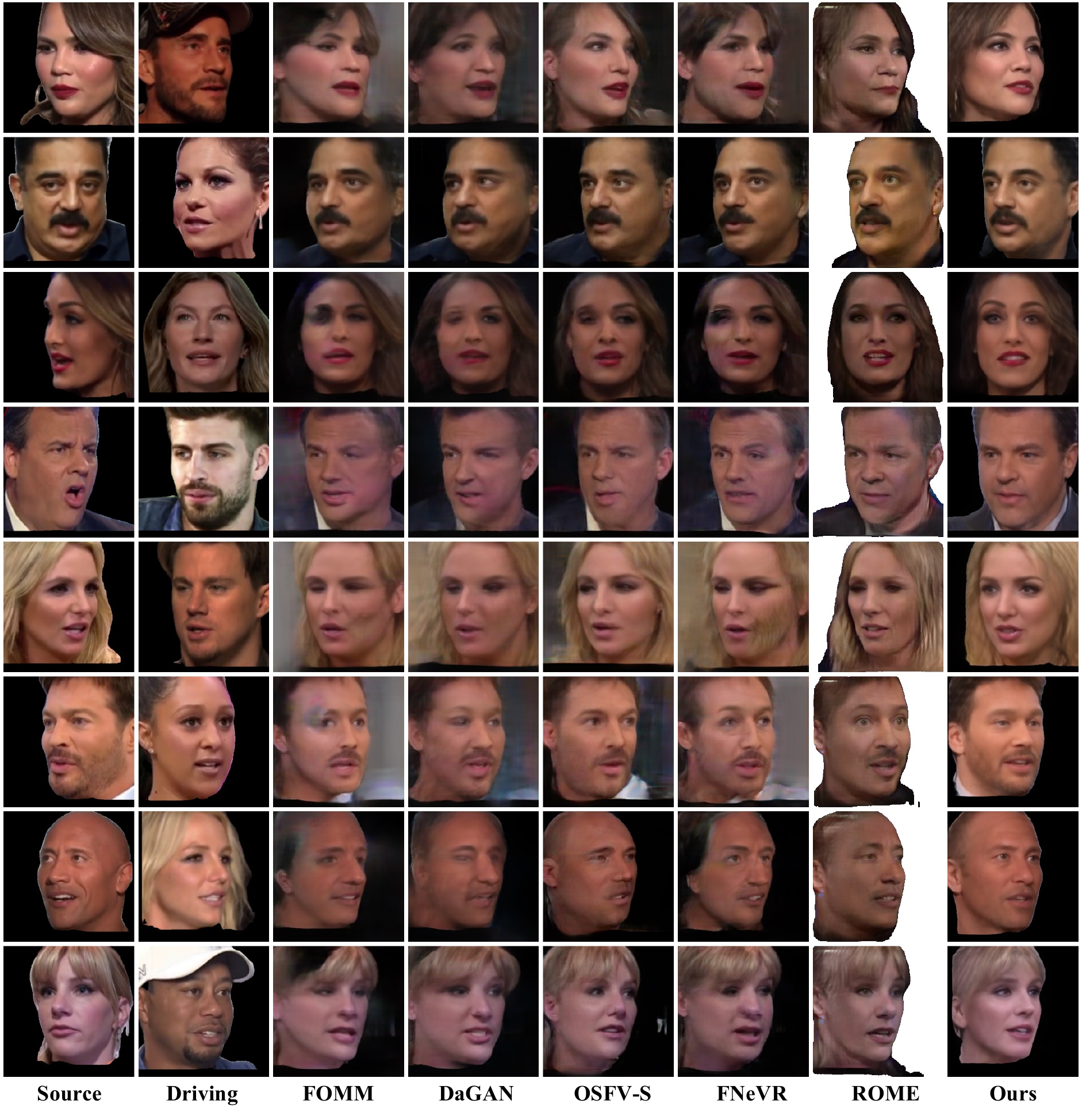}

\caption{
\textbf{Qualitative comparisons of cross-identity reenactment on the VoxCeleb2 dataset\cite{Chung18b}. }
  }
  \label{fig:sup_vox2}
  \vspace{-4mm}
\end{figure*}
\begin{figure*}[ht]
  \centering
  \includegraphics[width=\linewidth]{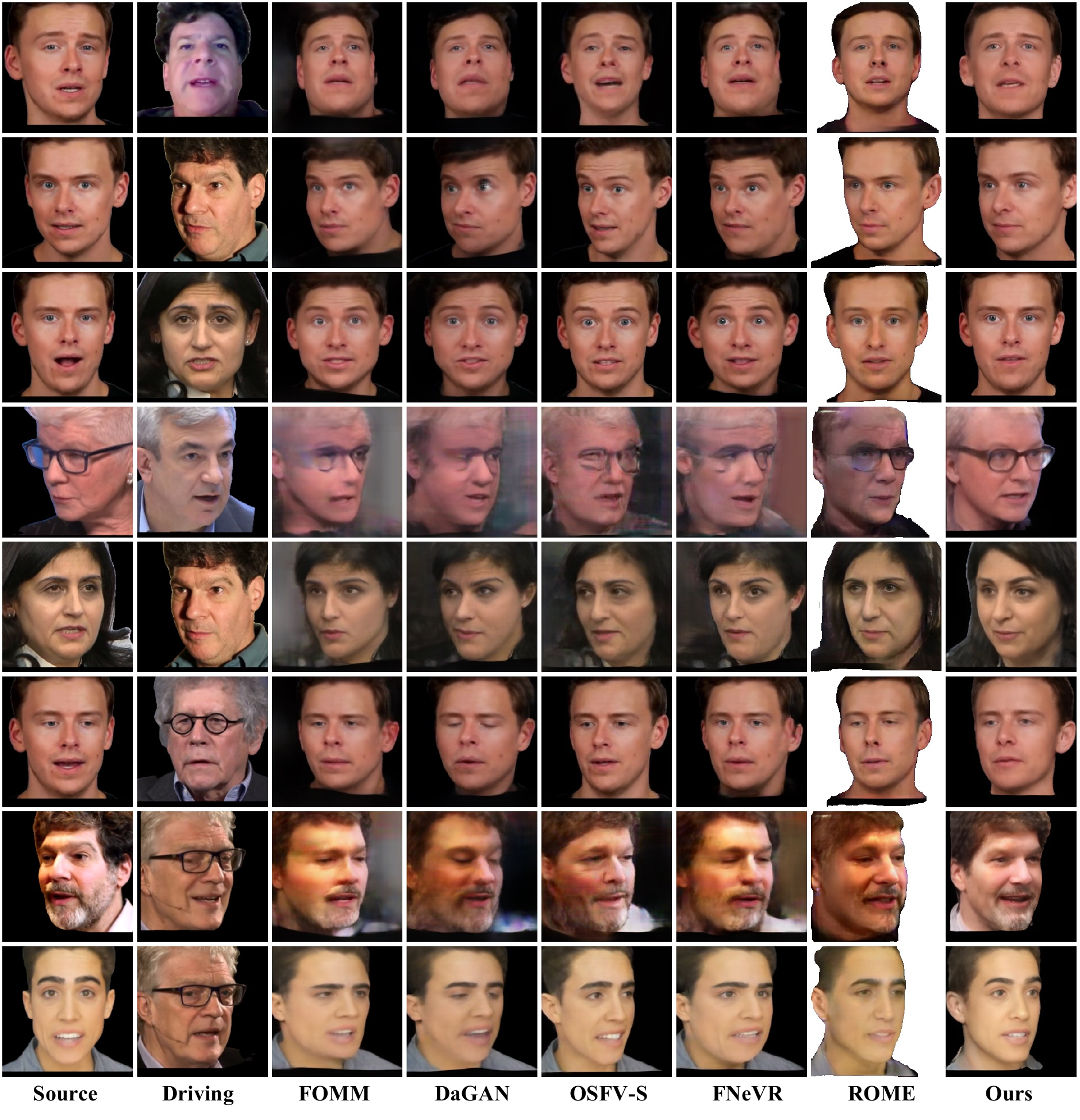}

\caption{
\textbf{Qualitative comparisons of cross-identity reenactment on the TalkingHead-1KH dataset\cite{wang2021facevid2vid}. }
  }
  \label{fig:sup_talk}
  \vspace{-4mm}
\end{figure*}

\subsection{Metrics}

\noindent\textbf{PSNR} is numerically related to the mean squared error (MSE) between the ground truth and the reconstructed image, it is used to measure the image reconstruction quality. 

\noindent\textbf{SSIM} measures the structural similarity between patches of the input images.
As a result, it is more robust to changes in the global illumination than PSNR.

\noindent\textbf{LPIPS\cite{zhang2018unreasonable}} 
calculates the cosine distances between the network features of the two images layer by layer and averages them to estimate the perceived distance of the generated image from the ground truth image.

\noindent\textbf{CSIM\cite{zhang2018unreasonable}} 
To evaluate the effectiveness of identity preservation, we compute the cosine similarity using embedded vectors created by the pre-trained face recognition model.

\noindent\textbf{AUCON} 
is used to calculate the ratio of the same facial action unit values between the generated images and the driving images.

\subsection{Ablation Study}

We use self-reenactment for qualitative comparison as the driving image serves as the ground truth, which will better exhibit identity preservation. 
As shown in the Fig.~\ref{fig:sup_ablation}, our proposed method can preserve the identity information from the source image quite well. 
However, when removing the multi-scale module, the identity information is largely change, the skin tone and texture cannot be maintained quite well, but the generated result can still mimic the driving motion. 
If we remove the camera-to-world transformation, the pose of the generated image cannot be changed precisely. 
We found that the rendered image is quite blurry and cannot guarantee large pose changes. 
As shown in the last row of Fig.~\ref{fig:sup_ablation}, entangling the pose with expression will lead to inaccurate expression, there is a problem with the eye and mouth movements in the generated image.

\subsection{Additional Qualitative Results}
% \subsection{More visual results}

In this section, we will show more qualitative results of cross-identity reenactment on different datasets. 
We show some representative results in this supplementary, Fig.~\ref{fig:sup_1} and Fig.~\ref{fig:sup_2} contains cross-identity reenactment on the VoxCeleb1 dataset. Fig.~\ref{fig:sup_vox2} and Fig.~\ref{fig:sup_talk} shows the cross-identity reenactment on the VoxCeleb2 and TalkingHead-1KH dataset, respectively. 

There is a diversity of disparities in the presented images (\eg, gender, face shape, skin color, beard, \etc).
Other methods have difficulty in maintaining the identity information of the source image under such disparities.
The facial shape of images generated with warping-based methods are easily affected by the driving image,
especially when the face shape difference is large. 
However, our method is able to maintain better identity information under all these conditions.\looseness=-1

% \noindent\textbf{Cross Identity Rendering}

% \input{Sup_Figs/sup_cross_id_1}
% \input{Sup_Figs/sup_cross_id_2}
% \input{Sup_Figs/sup_vox2}
% \input{Sup_Figs/sup_talk1kh}
% \input{Figs/sup_cross_id_3}

\begin{figure*}[ht]
%   \vspace{-8mm}
  \centering
  \includegraphics[width=\linewidth]{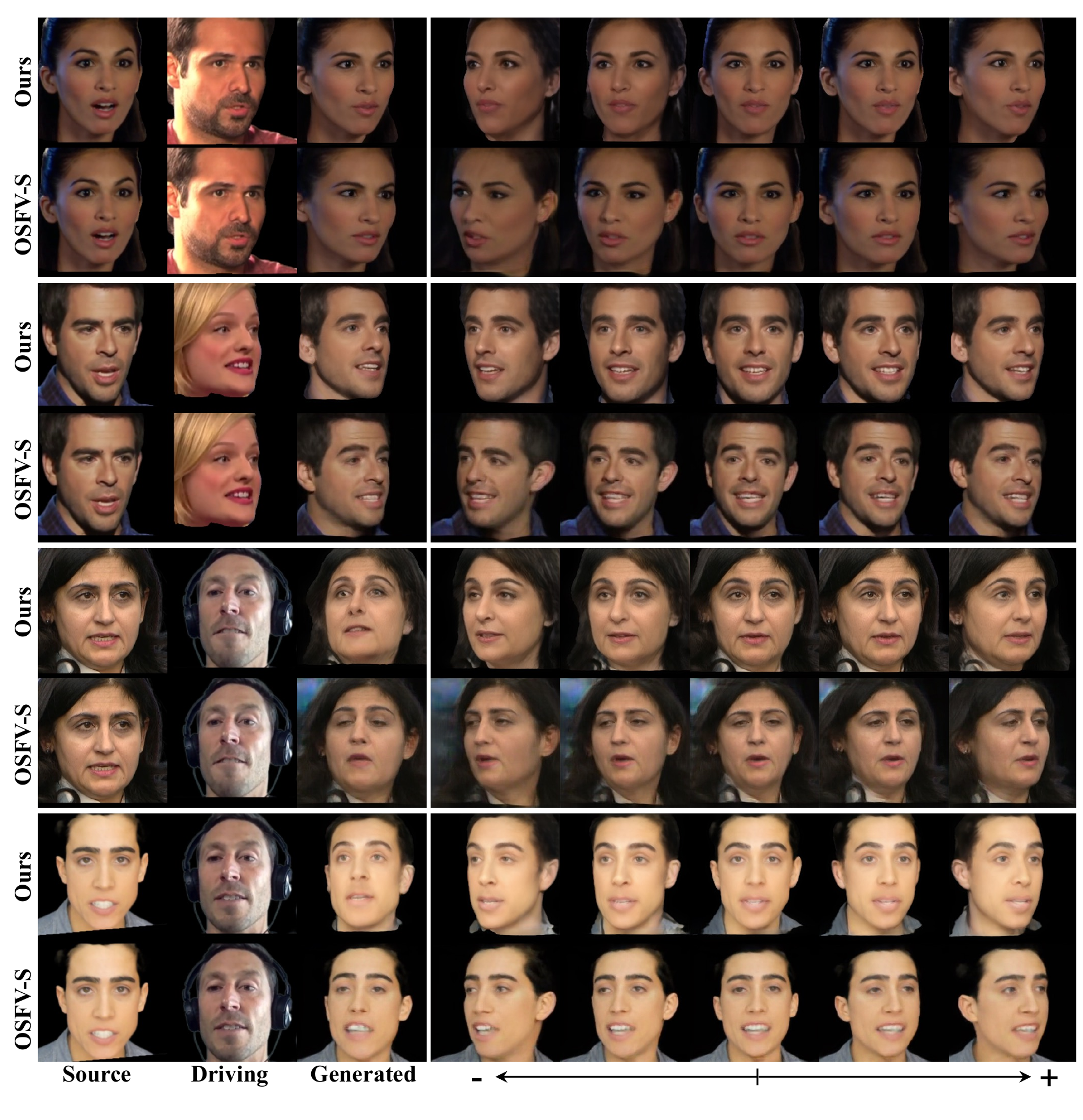}

% \vspace{-7mm}
\caption{
\textbf{Extrapolation to steep yaw angles.}
The first two rows contains the generated images from VoxCeleb1 dataset, while the last two rows are from TalkingHead-1KH datast. 
  }
  \label{fig:sup_fv_yaw}
%   \vspace{-2mm}
\end{figure*}

\begin{figure*}[ht]
%   \vspace{-8mm}
  \centering
  \includegraphics[width=\linewidth]{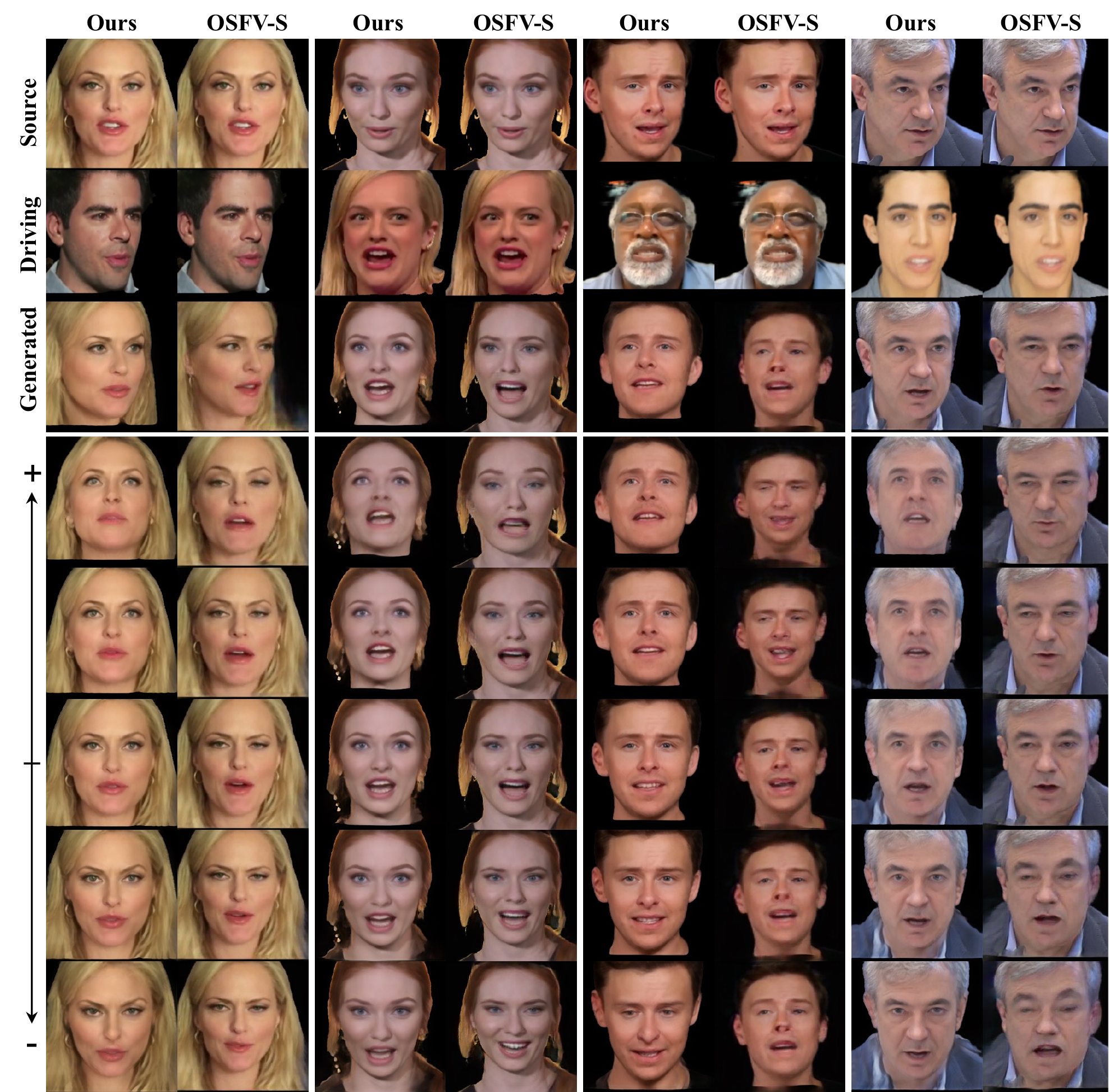}

% \vspace{-5mm}
\caption{
\textbf{Extrapolation to steep pitch angles.}
The first two columns contains the generated images from VoxCeleb1 dataset, while the last two columns are from TalkingHead-1KH datast.   }
  \label{fig:sup_fv_pitch}
%   \vspace{-2mm}
\end{figure*}

\subsection{Free-view Rendering}

Fig.~\ref{fig:sup_fv_yaw} and Fig.~\ref{fig:sup_fv_pitch} provides a visual comparison of our method against OSFV-S~\cite{wang2021facevid2vid} for generating views from steep camera poses, the first two columns/rows contains the generated results of VoxCeleb1~\cite{Nagrani17} and the other two columns/rows contains the results of TalkingHead-1KH~\cite{wang2021facevid2vid}.  
We observe that front-facing photos make up most of the commonly used talking-head datasets. 
Severe yaw and pitch angles are rarely shown in photographs, and neither is extreme yaw angles.
Yet, reasonable extrapolation to different poses is a critical trait for real-world talking head synthesis.\looseness=-1 

As shown in Fig.~\ref{fig:sup_fv_yaw}, our method is able to cope with different angles, showing satisfactory results with different view directions. 
However, OSFV struggles to maintain the expression-consistency when shifting the view angles, which is undesired in generating talking-heads in the real-world.
At the same time, when shifting the yaw angle, our method is able to rotate more accurately without changing other angles. 
Meanwhile, as shown in Fig.~\ref{fig:sup_fv_pitch}, OSFV-S has difficulty in rotating the pitch angle. 
Even though we change the viewing angle considerably, there was no significant change in the visual effect of the generated images. 
This phenomenon aligns with our observation that the variation in pitch angle in the dataset is small.

\subsection{Failure Cases}

As shown in Fig.~\ref{fig:bad}, our method generates results with degraded quality when there are occlusions in the images, such as microphones, and sunglasses. 
Also, our method cannot faithfully preserve the identity information under extreme pose changes.

\clearpage

\clearpage
%%%%%%%%% REFERENCES
% {\small
% \bibliographystyle{ieee_fullname}
% \bibliography{egbib}
% }

\end{document}